\documentclass[nonacm,acmtog]{acmart}

\PassOptionsToPackage{colorlinks,allcolors=black}{hyperref}
\usepackage{hyperref}

\usepackage{booktabs} 

\usepackage[ruled]{algorithm2e} 

\SetAlFnt{\small}
\SetAlCapFnt{\small}
\SetAlCapNameFnt{\small}
\SetAlCapHSkip{0pt}

\usepackage{booktabs} 
\usepackage{multirow} 
\usepackage{algorithmicx}
\usepackage{colortbl}

\begin{document}
\title{Textured 3D Regenerative Morphing with 3D Diffusion Prior}

\author{Songlin Yang}
\affiliation{%
 \institution{S-Lab, Nanyang Technological University}
 \country{Singapore}
}
\email{songlin.yang@ntu.edu.sg}
\author{Yushi Lan}
\affiliation{%
 \institution{S-Lab, Nanyang Technological University}
 \country{Singapore}
}
\email{yushi001@e.ntu.edu.sg}
\author{Honghua Chen}
\affiliation{%
 \institution{S-Lab, Nanyang Technological University}
 \country{Singapore}
}
\email{honghua.chen@ntu.edu.sg}
\author{Xingang Pan}
\affiliation{%
 \institution{S-Lab, Nanyang Technological University}
 \country{Singapore}
}
\email{xingang.pan@ntu.edu.sg}


\begin{abstract}

Textured 3D morphing creates \textit{smooth} and \textit{plausible} interpolation sequences between two 3D objects, focusing on transitions in both shape and texture. This is important for creative applications like visual effects in filmmaking. Previous methods rely on establishing point-to-point correspondences and determining smooth deformation trajectories, which inherently restrict them to shape-only morphing on untextured, topologically aligned datasets. This restriction leads to labor-intensive preprocessing and poor generalization. To overcome these challenges, we propose a method for 3D regenerative morphing using a 3D diffusion prior. Unlike previous methods that depend on explicit correspondences and deformations, our method eliminates the additional need for obtaining correspondence and uses the 3D diffusion prior to generate morphing. Specifically, we introduce a 3D diffusion model and interpolate the source and target information at three levels: initial noise, model parameters, and condition features. We then explore an \textbf{\textit{Attention Fusion}} strategy to generate more smooth morphing sequences. To further improve the plausibility of semantic interpolation and the generated 3D surfaces, we propose two strategies: \textbf{\textit{(a) Token Reordering}}, where we match approximate tokens based on semantic analysis to guide implicit correspondences in the denoising process of the diffusion model, and \textbf{\textit{(b) Low-Frequency Enhancement}}, where we enhance low-frequency signals in the tokens to improve the quality of generated surfaces. Experimental results show that our method achieves superior smoothness and plausibility in 3D morphing across diverse cross-category object pairs, offering a novel regenerative method for 3D morphing with textured representations.
  
\end{abstract}

%
%


%
%


\begin{teaserfigure}
\centering
  \includegraphics[width=1.0\textwidth]{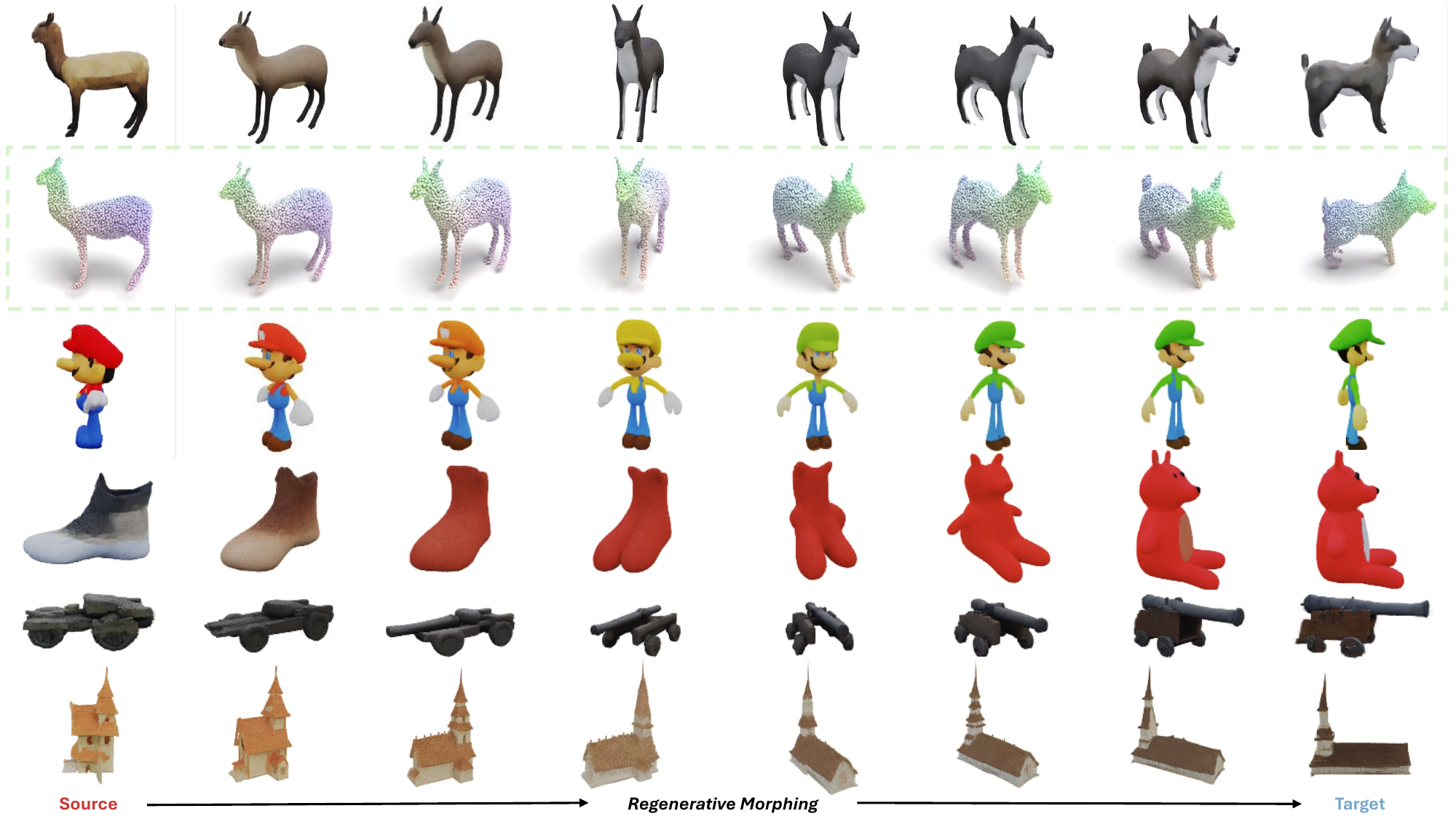}
  \caption{The textured 3D morphing sequences regenerated using our method with 3D diffusion prior. Our method requires no category-specific alignment data for model training and no labor-intensive preprocessing for explicit correspondence, enabling smooth and plausible morphing across diverse cross-category 3D object pairs. \textit{ \textbf{More video results can be found \href{https://songlin1998.github.io/Textured-3D-Morphing/}{\textcolor{red}{here}}}}.}
  \label{fig:teaser}
\end{teaserfigure}

\maketitle

\section{Introduction}

Morphing~\cite{lin20242d} generates an interpolation sequence between a source and a target, requiring smooth and plausible transitions. This fundamental technique is crucial for creative applications like visual effects in film and media. Depending on the data type, morphing is categorized into image morphing~\cite{shechtman2010regenerative,zhang2024diffmorpher} and 3D morphing~\cite{tsai2022multiview,kim2024meshup,geng2024birthdeathrose}. 
Compared to image morphing, 3D morphing aligns more naturally with visual effects production, where objects and transformations inherently operate in 3D space.
However, 3D morphing is more challenging, requiring the interpolation of 3D objects holistically (i.e., image morphing can be viewed as a special case of 3D morphing from a specific viewpoint). Our work addresses the task of textured 3D morphing, which generates a sequence for two textured 3D representations, aiming for smooth and plausible transitions in shape and texture, as shown in Fig.~\ref{fig:teaser}.

Previous 3D morphing methods mainly focused on morphing shapes, which can be summarized in two steps: \textit{first}, establishing correspondence~\cite{deng2023se} between the 3D representations of the source and target objects, and \textit{second}, determining smooth and plausible deformation trajectories between corresponding 3D points~\cite{eisenberger2021neuromorph}. Following these steps, previous methods blend the two 3D representations with respective weights to obtain a sequence of interpolated 3D representations.



Due to the scarcity of topologically aligned and textured 3D datasets, previous 3D morphing methods have mainly focused on in-domain untextured datasets, such as FAUST~\cite{bogo2014faust} (human shapes) and Shrec'20~\cite{dyke2020shrec} (quadruped animals). As a result, these methods are limited to \textit{\textbf{shape-only morphing}} and face two key \textit{\textbf{generalization}} challenges: \textit{(a) Labor-Intensive Preprocessing}: Morphing new in-domain 3D data with these methods~\cite{eisenberger2021neuromorph,zhan2024charactermixer} requires domain-specific alignment with the training datasets through tedious registration~\cite{sun2024srif} and matching~\cite{zhu2024densematcher} steps. \textit{(b) Limited Morphing Capacity}: The methods~\cite{aydinlilar2021part,vyas2021latent,eisenberger2021neuromorph,zhan2024charactermixer} suffer from limited object diversity and small datasets, leading to ambiguous and implausible interpolations.


The limitations of previous methods inspire us to consider two critical questions: \textit{(a) Is explicit point-to-point correspondence truly necessary? (b) Can we enhance the generalization capability of textured 3D morphing via a generic generative prior?}


For \textit{\textbf{correspondence}}, obtaining dense correspondences between textured 3D representations across categories remains underexplored~\cite{zhu2024densematcher} despite being foundational to previous methods, while explicit constraints can instead limit the creativity of morphing. Therefore, a promising method is using labor-saving implicit correspondences~\cite{lan2022ddf_ijcv,yang2024learning} to guide the morphing. For instance, morphing could be formulated as an optimization problem, allowing correspondences to emerge automatically~\cite{tsai2022multiview}. Alternatively, attention mechanisms in generative models could enable automatic alignment during object blending~\cite{zhang2024diffmorpher,he2024aid,shen2024dreammover}. 

For\textit{ \textbf{generative priors}}, recent advancements in diffusion-based generation models offer two strategies for 3D morphing: using 3D diffusion models directly~\cite{lan2024gaussiananything, chen20243dtopia, xiang2024structured}, or enhancing 2D models with 3D priors. However, those hybrid 2D-3D methods~\cite{rombach2022high, kim2024meshup,poole2022dreamfusion,haque2023instruct} face significant challenges: 2D models lack a holistic 3D knowledge, and optimizing the 2D-3D mapping is difficult, with no guarantee that the morphing process will maintain 3D consistency across viewpoints. Therefore, directly using a 3D diffusion model to regenerate the interpolated 3D representations (i.e., regenerative morphing) enables authentic 3D morphing and has the potential to be scaled up by using a more state-of-the-art 3D generation model. 


Building on the above analysis, we adopt a generic 3D diffusion prior that leverages its implicit correspondence and 3D generation capabilities to blend source and target information, enabling the regeneration of interpolated textured 3D representations.

Specifically, we first introduce a 3D diffusion model~\cite{lan2024gaussiananything} and interpolate the information from the source and target at three levels: initial noises, model parameters, and condition features. The 3D representation for each interpolation is then regenerated using the 3D generation model. To improve the 3D diffusion model's ability to generate smooth morphing sequences, we explore an \textit{\textbf{Attention Fusion}} strategy. However, fusing different information weakened the model's denoising capability, and aggressively applying Attention Fusion to all denoising steps for smoother morphing resulted in implausible outcomes. Therefore, to improve the plausibility of semantic interpolation and the generated 3D surfaces, we propose two strategies: \textit{\textbf{(a) Token Reordering}}: After semantic analysis, we identify semantic correspondences in diffusion tokens and propose matching approximate tokens before attention computation to better guide implicit correspondences in the diffusion space. \textit{\textbf{(b) Low-Frequency Enhancement}}: Frequency-domain analysis reveals that boosting low-frequency signals improves the surface quality of the 3D diffusion model. Thus, we enhance low-frequency signals at key time steps to preserve the model’s ability to generate 3D surfaces.

Our contributions can be summarized as follows: (a) To the best of our knowledge, we are the first to use a generic 3D diffusion prior for morphing textured 3D representations, enabling 3D morphing without explicit correspondences. (b) We analyze the merging of source and target information during morphing with a 3D diffusion prior from semantic and frequency perspectives, proposing Token Reordering and Low-Frequency Enhancement to improve smoothness and plausibility. (c) Extensive experimental results demonstrate that our method achieves superior smoothness and plausibility in performing 3D morphing across diverse cross-category object pairs.

\vspace{-0.3cm}
\section{Related Work}

\subsection{3D Morphing}

Previous 3D morphing methods focus on shape-only correspondence~\cite{tam2012registration}, and realize 3D morphing through interpolation or deformation between corresponding 3D primitives (e.g., points, vertices, and faces). They can be divided into \textit{(a) Axiomatic Methods:} They tend to rely on sparse landmarks~\cite{kim2011blended,edelstein2019enigma} or use functional maps~\cite{ovsjanikov2012functional} to address under-constrained mapping spaces, such as MapTree~\cite{ren2020maptree} and SmoothShells~\cite{eisenberger2020smooth}. Energy-minimizing functions~\cite{sorkine2007rigid,sorkine2004laplacian} and skinning methods~\cite{fulton2019latent,jacobson2014skinning} enable deformation control, while optimal transportation~\cite{solomon2015convolutional,tsai2022multiview} approximates shape correspondence. \textit{(b) Learning-Based Methods:} This category of methods can be categorized into introducing other generative priors and using aligned data. SATR~\cite{abdelreheem2023satr} introduces the semantic labels from multi-view images. NSSM~\cite{morreale2024neural} uses Dinov2~\cite{oquab2023dinov2} for sparse landmarks and SRIF~\cite{sun2024srif} introduces large vision models~\cite{zhang2024diffmorpher} for semantic shape registration and morphing. A recent class of works leverages text prompts as user inputs for driving a deformation towards an arbitrary textual prompt~\cite{gao2023textdeformer,mohammad2022clip,michel2022text2mesh}, but CLIP~\cite{radford2021learning} objective lacks a full understanding of object details. For data prior, category-specific training~\cite{yumer2015semantic} on a topology-aligned dataset such as NeuroMorph~\cite{eisenberger2021neuromorph} is also prevailing in learning-based 3D shape analysis. 

These methods focus on shape-only morphing or rely on rigging annotations, requiring extra effort due to 2D-3D mismatches. In contrast, our work enables textured 3D morphing without learning correspondence or deformation, using a 3D diffusion model to regenerate interpolated representations, addressing prior challenges and offering a new direction for 3D morphing research.

\vspace{-0.4cm}
\subsection{Image Morphing}

Image morphing is a long-standing challenge in computer vision and graphics~\cite{aloraibi2023image, wolberg1998image, zope2017survey}. Traditional methods~\cite{beier2023feature, bhatt2011comparative, darabi2012image, liao2014automating, shechtman2010regenerative} use feature-based warping and blending to create smooth transitions but struggle to generate new content, often leading to artifacts. Data-driven methods~\cite{averbuch2016smooth, fish2020image} leverage large single-class datasets to achieve smoother results but are limited in cross-domain or personalized applications due to their reliance on specific data. The methods like DiffMorpher~\cite{zhang2024diffmorpher} and AID~\cite{he2024aid} address this by utilizing pre-trained diffusion models on diverse datasets, enabling flexible morphing across a wide range of object categories. \textit{Inputting multi-view images to image morphing methods enables 3D-aware morphing.}

\vspace{-0.4cm}
\subsection{3D Diffusion Model}

Generative 3D priors fall into two types: native 3D diffusion models and 3D-aware diffusion models. Due to the scarcity of 3D data, methods leveraging 2D priors to generate 3D or multi-view content have been proposed. For instance, Score Distillation Sampling~\cite{poole2022dreamfusion} distills 3D information from a 2D diffusion model. 3D-aware generation can be divided into two steps: multi-view image generation~\cite{shi2023mvdream} followed by feed-forward 3D reconstruction~\cite{xu2024instantmesh}. However, these methods inherently lack a 3D latent space. To address this, native 3D diffusion models~\cite{zhang20233dshape2vecset,vahdat2022lion,lan2025ln3diff} that encode and learn 3D representations have been introduced. These models typically consist of two steps: training a VAE to encode 3D data and training a diffusion model based on corresponding latent codes. We use Gaussian Anything~\cite{lan2024gaussiananything} as the 3D diffusion prior, encoding 3D information in a point cloud latent space.

\begin{figure*}[t]
    \centering
    \includegraphics[width=\textwidth]{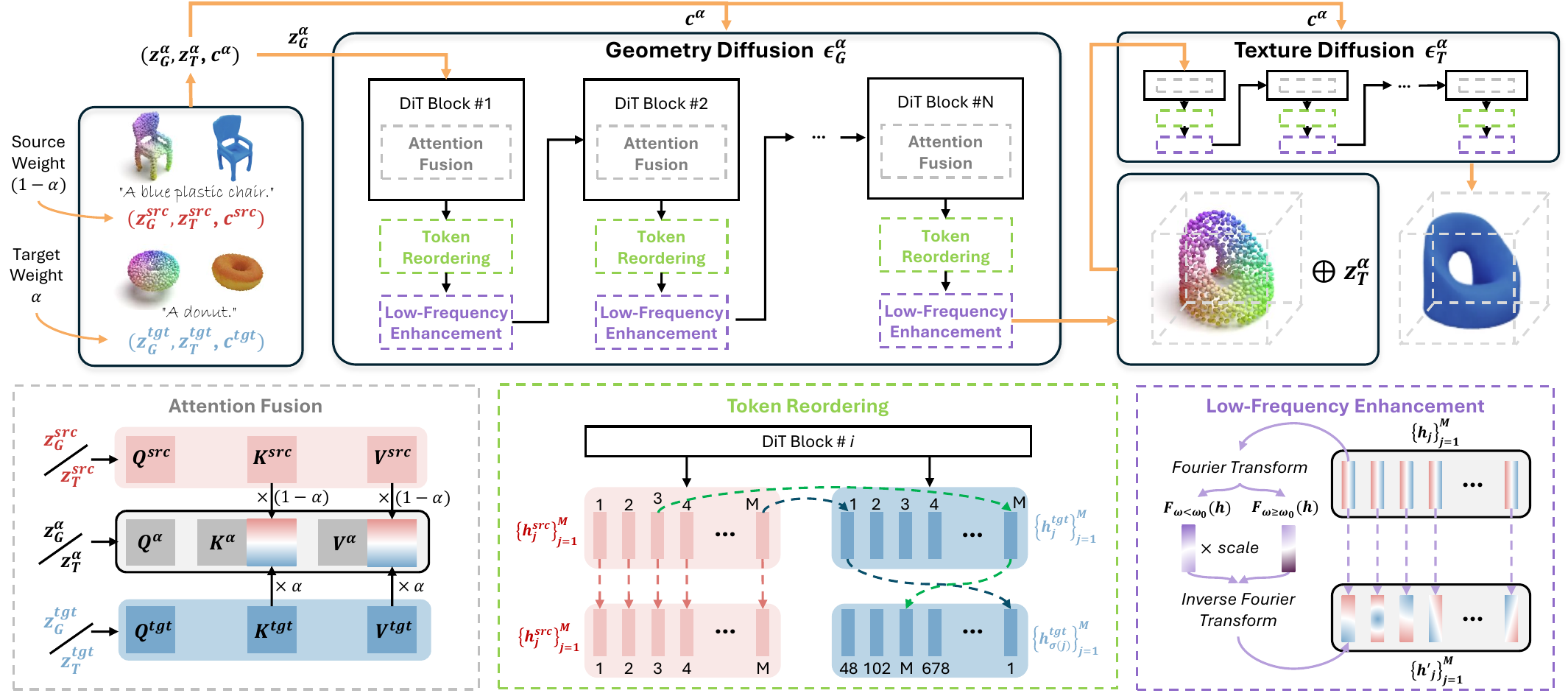}
    \vspace{-0.6cm}
    \caption{The framework of our method. The 3D diffusion prior is a two-stage (geometry \& texture) generation model. Beyond basic interpolation, Attention Fusion  is explored to improve smoothness, while Token Reordering  and Low-Frequency Enhancement are proposed to improve plausibility.}
    \label{fig: framework}
    \vspace{-0.1cm}
\end{figure*}

\section{Method}

The pipeline for 3D regenerative morphing based on the 3D diffusion prior, as shown in Fig.\ref{fig: framework}, consists of three main steps: \textit{\textbf{Basic Interpolation}}, where essential information is interpolated (Sec.\ref{method: basic}); \textit{\textbf{Smoothness Improvement}}, achieved through an Attention Fusion mechanism (Sec.\ref{method: attention fusion}); and \textit{\textbf{Plausibility Improvement}}, which involves two strategies, Token Reordering (Sec.\ref{method: token reordering}) and Low-Frequency Enhancement (Sec.~\ref{method: low-frequency enhancement}). 

\subsection{Preliminary}
\label{method: preliminary}

\subsubsection{3D Diffusion Model}

We select Gaussian Anything~\cite{lan2024gaussiananything} as our 3D diffusion prior, a two-stage native 3D diffusion model with a structured latent representation (i.e., point cloud) and the DiT~\cite{peebles2023scalable} backbone. It consists of a geometry generation model $\epsilon_G$ and a texture generation model $\epsilon_T$. In the first stage, the model $\epsilon_G$ takes a Gaussian initial noise $\mathbf{z}_{\mathit{G}}$ and textual conditioning information $\mathbf{c}$ as inputs, to generate a structured point cloud representation $\mathbf{x}_{\mathit{point\text{-}cloud}} = \epsilon_G(\mathbf{z}_{\mathit{G}}, \mathbf{c})$. In the second stage, $\mathbf{x}_{\mathit{point\text{-}cloud}}$ is added to a Gaussian initial noise $\mathbf{z}_{\mathit{T}}$, and the resulting variable is denoised by $\epsilon_T$ with condition $\mathbf{c}$ to obtain the final texture feature $\mathbf{x}_{\mathit{feature}} = \epsilon_T(\mathbf{x}_{\mathit{point\text{-}cloud}} + \mathbf{z}_{\mathit{T}}, \mathbf{c})$. Finally, $\mathbf{x}_{\mathit{point\text{-}cloud}}$ and $\mathbf{x}_{\mathit{feature}}$ are fed into a pre-trained decoder $\mathcal{D}$ to produce the final 3D Gaussian representation $\mathbf{x}_{\mathit{3D}} = \mathcal{D}(\mathbf{x}_{\mathit{point\text{-}cloud}}, \mathbf{x}_{\mathit{feature}})$, which can be rendered as multi-view images. We denote the token sequences processed between DiT blocks as $\{ h_j \}_{j=1}^{M}$, where $M$ is the length of token sequence.

\subsubsection{Attention} The attention~\cite{vaswani2017attention} mechanism is an important component for the current text-driven diffusion models~\cite{rombach2022high,peebles2023scalable,saharia2022photorealistic,chen20243dtopia,xiang2024structured}, especially cross-attention and self-attention. Given a latent variable $z \in \mathbb{R}^{d_z}$, a text condition $ c \in \mathbb{R}^{d_c} $, and the attention layer with matrices $ W_Q \in \mathbb{R}^{d_z \times d_q} $,$ W_K \in \mathbb{R}^{d_c \times d_k} $, and $ W_V \in \mathbb{R}^{d_c \times d_v} $, the cross-attention is computed as: 
\vspace{-0.2cm}
\begin{equation}
    A(z, c) = \text{Attn}(Q, K, V) = \text{softmax}\left(\frac{QK^\top}{\sqrt{d_k}}\right) V,
    \label{eqn: attention}
\end{equation}

\noindent
where $Q=W_Q^\top z, K=W_K^\top c, V=W_V^\top c$. Self-attention is a special case of cross-attention and can be computed with $A(z, z)$.

\subsection{Basic Interpolation}
\label{method: basic}

\subsubsection{Implementation}

The basic interpolation has three levels, with source and target weights given by $(1 - \alpha)$ and $ \alpha $, respectively, where $\alpha = 0 $ generates the source $\mathbf{x}_{\mathit{3D}}^{\mathit{src}} $, and $\alpha = 1$ generates the target $ \mathbf{x}_{\mathit{3D}}^{\mathit{tgt}}$.

\textit{(a) Initial Noises}: The textured 3D representations $\mathbf{x}_{\mathit{3D}}^{\mathit{src}}$ and $\mathbf{x}_{\mathit{3D}}^{\mathit{tgt}}$, are inverted via diffusion inversion~\cite{lipman2022flow} to obtain their respective input noises, $[ \mathbf{z}_{\mathit{T}}^{\mathit{src}}, \mathbf{z}_{\mathit{G}}^{\mathit{src}} ]$ and $[ \mathbf{z}_{\mathit{T}}^{\mathit{tgt}}, \mathbf{z}_{\mathit{G}}^{\mathit{tgt}} ]$. To ensure Gaussian noise properties are preserved, spherical linear interpolation~\cite{samuel2024norm} is applied to these noises to generate the interpolated noises, $[ \mathbf{z}_{\mathit{T}}^{\alpha}, \mathbf{z}_{\mathit{G}}^{\alpha} ]$.

\textit{(b) Model Parameters}: Given $ \mathbf{x}_{\mathit{3D}}^{\mathit{src}}$ and $ \mathbf{x}_{\mathit{3D}}^{\mathit{tgt}} $, we fine-tune the model using LoRA (Low-Rank Adaptation)~\cite{hu2021lora} to obtain two sets of LoRA parameters. These parameters are then linearly interpolated and fused to obtain the morphing models $ \epsilon_G^\alpha $ and $ \epsilon_T^\alpha $. 

\textit{(c) Condition Features}: To enhance semantic consistency, the text prompts of the source and target 3D representations are encoded via a CLIP~\cite{radford2021learning} encoder into $ \mathbf{c}^{\mathit{src}}$ and $ \mathbf{c}^{\mathit{tgt}}$, which are linearly interpolated to produce $\mathbf{c}^{\alpha}$.

\subsubsection{Problems}

Basic interpolation realizes the blending of basic information, but there are still the following problems:

\textit{(a) Abrupt Changes (Smoothness)}: Nonlinear multi-step denoising in diffusion models introduces variability in noise-to-data mapping, while the CLIP encoder’s space lacks guaranteed semantic smoothness, leading to abrupt changes (yellow-highlighted part in Fig.~\ref{fig: ablation}).

\textit{(b) Artifacts (Plausibility)}: Misalignment between conditioning and diffusion spaces disrupts learned mappings, causing artifacts like structure collapse or degraded surface quality (red and yellow-highlighted parts in Fig.~\ref{fig: ablation}).

 To address these, we investigate \textit{\textbf{Attention Fusion}} (Sec.~\ref{method: attention fusion}) for smoothness and \textit{\textbf{Token Reordering}} (Sec.~\ref{method: token reordering}) with \textit{\textbf{Low-Frequency Enhancement}} (Sec.~\ref{method: low-frequency enhancement}) for plausibility.

\subsection{Attention Fusion}
\label{method: attention fusion}


The fusion of attention has been proven effective in improving morphing smoothness in 2D diffusion-based image morphing. However, DiffMorpher~\cite{zhang2024diffmorpher} only explored self-attention interpolation and did not consider the smoothness of semantic features in conditioning. AID~\cite{he2024aid} did not address the differences between the original and LoRA models, neglecting conditioning feature interpolation with accurate inversion. Our method combines self-attention and cross-attention fusion, using unified attention features from fine-tuned models to enhance smoothness while ensuring plausible generation.

Specifically, as shown in Fig.~\ref{fig: framework}, we first feed $z^{src}$, $z^{tgt}$, and $z^\alpha$ into blocks of the morphing model $ \epsilon^\alpha $ to obtain three sets of ($Q^{src}$, $K^{src}$,$V^{src}$), ($Q^{tgt}$, $K^{tgt}$,$V^{tgt}$), and ($Q^\alpha$, $K^\alpha$,$V^\alpha$). Then, based on Eq.~(\ref{eqn: attention}), denoting concatenation as $[\cdot,\cdot]$, we obtain the fused attention by:

\vspace{-0.6cm}
\begin{multline}
    Fused\text{-}Attn(Q^\alpha, K^\alpha, V^\alpha) = \\
    Attn\left(Q^\alpha, \left[(1-\alpha)K^{\text{src}} + \alpha K^{\text{tgt}}, K^\alpha\right], \left[(1-\alpha)V^{\text{src}} + \alpha V^{\text{tgt}}, V^\alpha\right]\right).
\end{multline}
\vspace{-0.6cm}

Constraining all attention calculations to the same model helps mitigate the quality degradation caused by attention fusion. However, applying attention fusion across different time steps improves smoothness only to a point, beyond which plausibility declines, leading to structural collapse and surface issues (See Fig.~\ref{fig: ablation}). \textit{This emphasizes the need to balance smoothness with plausibility.}

\vspace{-0.3cm}
\subsection{Token Reordering}
\label{method: token reordering}
\subsubsection{Motivation} 3D objects are tokenized into sequences $\{ h_j \}_{j=1}^{M}$, with each token $h_j$ representing a 3D point. The DiT block’s attention modules and Attention Fusion guide the blending of source and target tokens using implicit correspondence during inference. However, relying solely on such implicit correspondence of attention mechanism to match the points in different 3D objects may lead the model to make semantically implausible connections (e.g., combining a chair leg with donut frosting). This vanilla application of the diffusion prior does not fully leverage its potential. Works like DIFT~\cite{tang2023emergent} show that 2D diffusion features/tokens can represent semantics~\cite{yu2024representation,zhang2025diff,hedlin2024unsupervised}. Similarly, we believe 3D diffusion features also capture 3D correspondences within the same object category and semantic correspondence across different object categories (e.g., the eyes of a dog and the eyes of a monkey). Therefore, \textit{why not pair points with similar semantics first, and then interpolate within this semantically plausible space?}

\begin{figure}[t]
    \centering
    \includegraphics[width=0.48\textwidth]{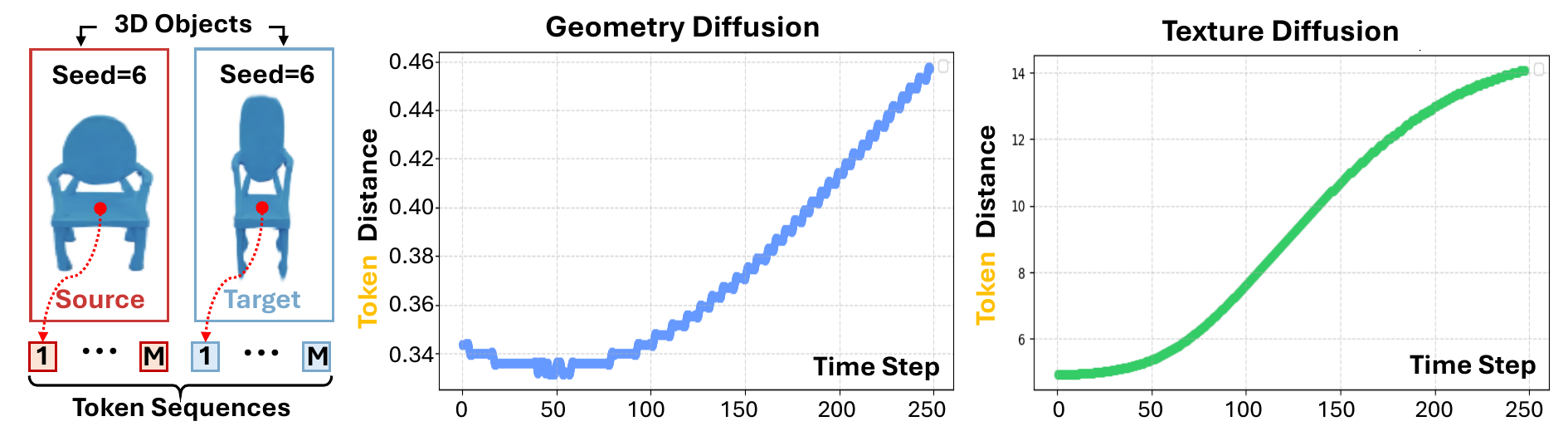}
    \vspace{-0.7cm}
    \caption{The token distances between tokens at the same position in the sequence. A 3D representation is scaled to generate a perfectly aligned version with the same random seeds, ensuring tokens at the identical sequence positions are semantically aligned. During denoising, the semantic distance between tokens at the identical positions increases.}
    \label{fig: reordering_1}
    \vspace{-0.2cm}
\end{figure}

\begin{figure}[t]
    \centering
    \includegraphics[width=0.47\textwidth]{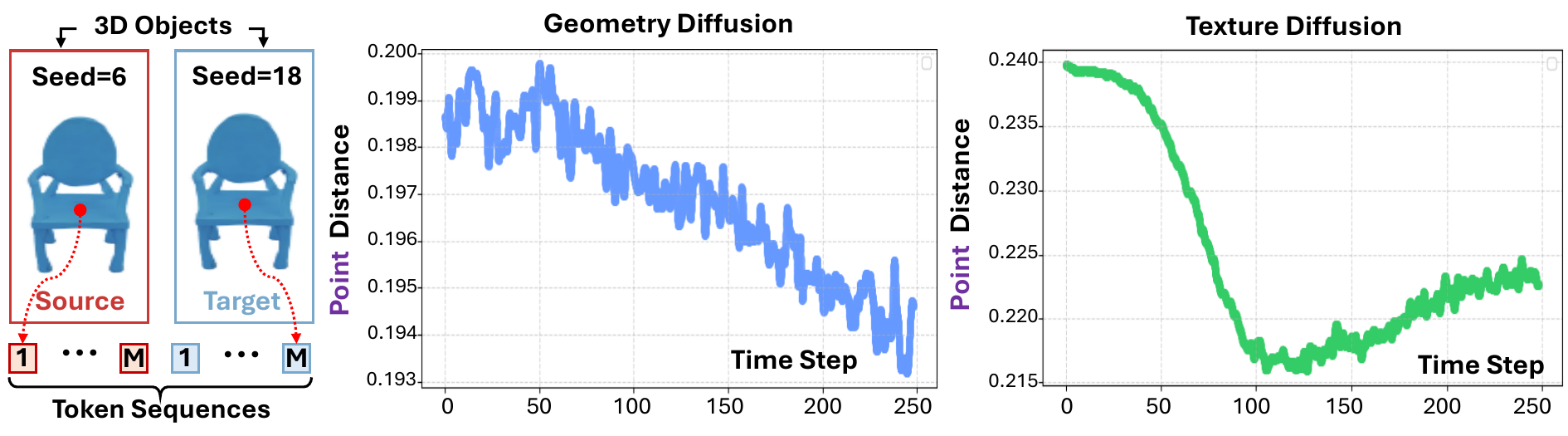}
    \vspace{-0.3cm}
    \caption{The point distances between token-distance-closest points. Different random seeds are used to generate varied initial noises, meaning tokens at the identical sequence positions do not correspond to the same 3D points. By extracting the two closest tokens and their corresponding points in the final point cloud, we computed the 3D distance between the paired points.}
    \label{fig: reordering_2}
    \vspace{-0.4cm}
\end{figure}

\subsubsection{Experimental Analysis} To validate our motivation, we tested the semantic correspondence of tokens on aligned data.

\textit{(a) Problems of Vanilla Attention Fusion.} The Fig.~\ref{fig: reordering_1} indicates that semantic alignment is lost during denoising, but vanilla Attention Fusion forcibly links these position-aligned tokens, which burdens the model and causes artifacts (See Fig.~\ref{fig: ablation}, third row).

\textit{(b) Existence of Semantic Correspondence.} As shown in Fig.\ref{fig: reordering_2}, the distance between points based on geometry token distance decreases with increasing time steps, indicating strengthened semantic correspondence. Conversely, the distance based on texture token distance first decreases and then increases, suggesting that the denoising process initially promotes semantic alignment, but later shifts towards learning texture details. This aligns with findings in 2D diffusion research\cite{yu2024representation}. Thus, by considering both geometry and texture denoising characteristics, we reorder tokens in the intermediate stages to better guide implicit correspondence for morphing.

\subsubsection{Implementation}
Based on these observations, we reorder the token sequences before passing the output of the $i$-th block to the $(i+1)$-th block, where $i \in \{1, 2, \dots, N\}$ and $N$ is the total number of blocks in the diffusion model. As shown in Fig.~\ref{fig: framework}, we reorder the source and target token sequences $\{ h_j^{\text{src}} \}_{j=1}^{M}$, $\{ h_j^{\text{tgt}} \}_{j=1}^{M}$ such that tokens with similar distances align at the same index, as follows:

\vspace{-0.2cm}
\begin{equation}
    \text{minimize} \sum_{j=1}^{M} \| \mathit{h}_j^{\text{src}} - \mathit{h}_{\sigma(j)}^{\text{tgt}} \|,
\end{equation}
\vspace{-0.1cm}

\noindent
where $M$ is the number of tokens, and $\sigma(j)$ denotes the index of the element in the target sequence that best corresponds to the $j$-th element in the source sequence. To further refine the generation, we set different reordering strategies depending on $\alpha$: For $\alpha \in [0, 0.5)$, the target token sequence is reordered based on the source token sequence. For $\alpha \in [0.5, 1]$, the source token sequence is reordered based on the target token sequence.

\subsection{Low-Frequency Enhancement}
\label{method: low-frequency enhancement}

\subsubsection{Motivation}

Due to the additional attention fusion operations, the misalignment between the condition space and the diffusion space is exacerbated. Excessive attention fusion at later time steps significantly degrades the model's denoising capability, with the deterioration becoming more pronounced as the time step increases (See Fig.~\ref{fig: ablation}, from the 4th to the 5th row). This raises the question:  \textit{what part of the 3D diffusion model is influenced by the morphing operations, leading to this decline in performance?} 

\begin{figure}[t]
    \centering
    \includegraphics[width=0.48\textwidth]{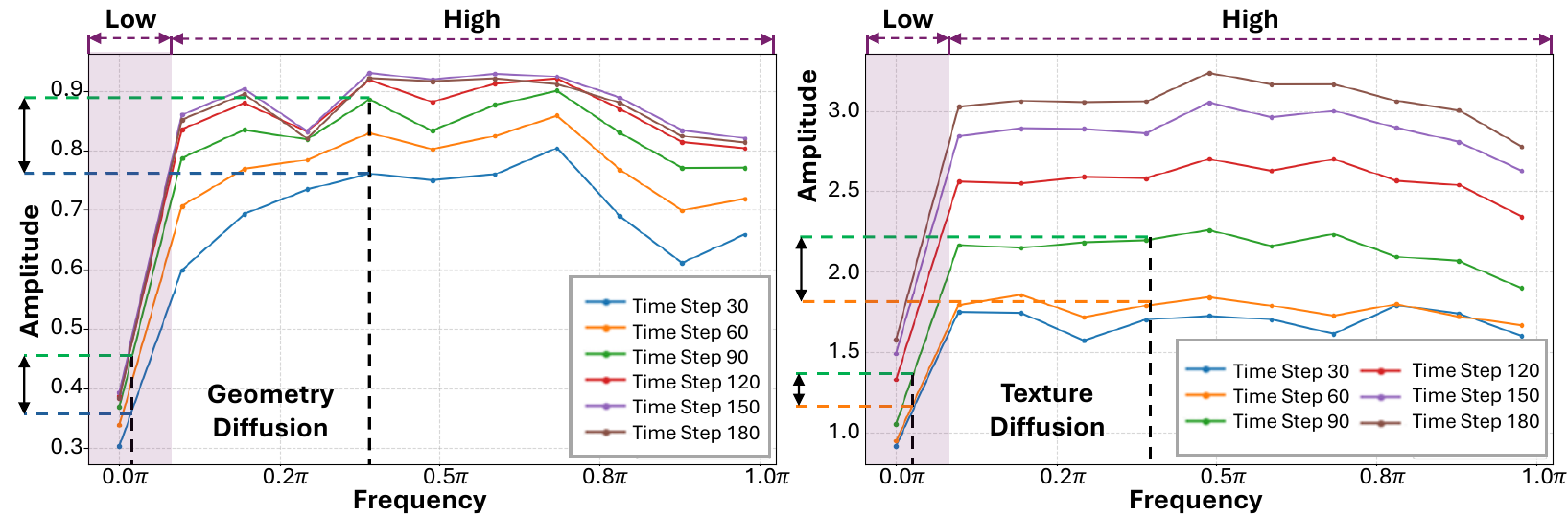}
    \vspace{-0.7cm}
    \caption{The changes of high and low frequency signals in the diffusion model during the denoising process. Visualizing signal amplitudes at different time steps reveals that low-frequency noise varies less than high-frequency noise, with smaller gaps across denoising time steps.}
    \label{fig: frequency}
    \vspace{-0.55cm}
\end{figure}

\subsubsection{Experimental Analysis}

The deterioration problem is analyzed in the frequency domain to address the significant visual differences between 3D objects, as shown in Fig.~\ref{fig: frequency}. In 3D generation, low-frequency noise controls the overall layout, while high-frequency noise governs surface details. Excessive amplification of high-frequency components during denoising can interfere with the low-frequency components, degrading overall quality. Therefore, enhancing low-frequency signals during denoising is crucial to prevent the model from overemphasizing high-frequency noise, thus improving the quality of 3D surface generation. Similar patterns appear in 2D diffusion studies~\cite{si2024freeu,wu2025freeinit}, with low frequencies tied to image layout and high frequencies to details.

\begin{figure*}[t]
    \centering
    \includegraphics[width=1.0\textwidth]{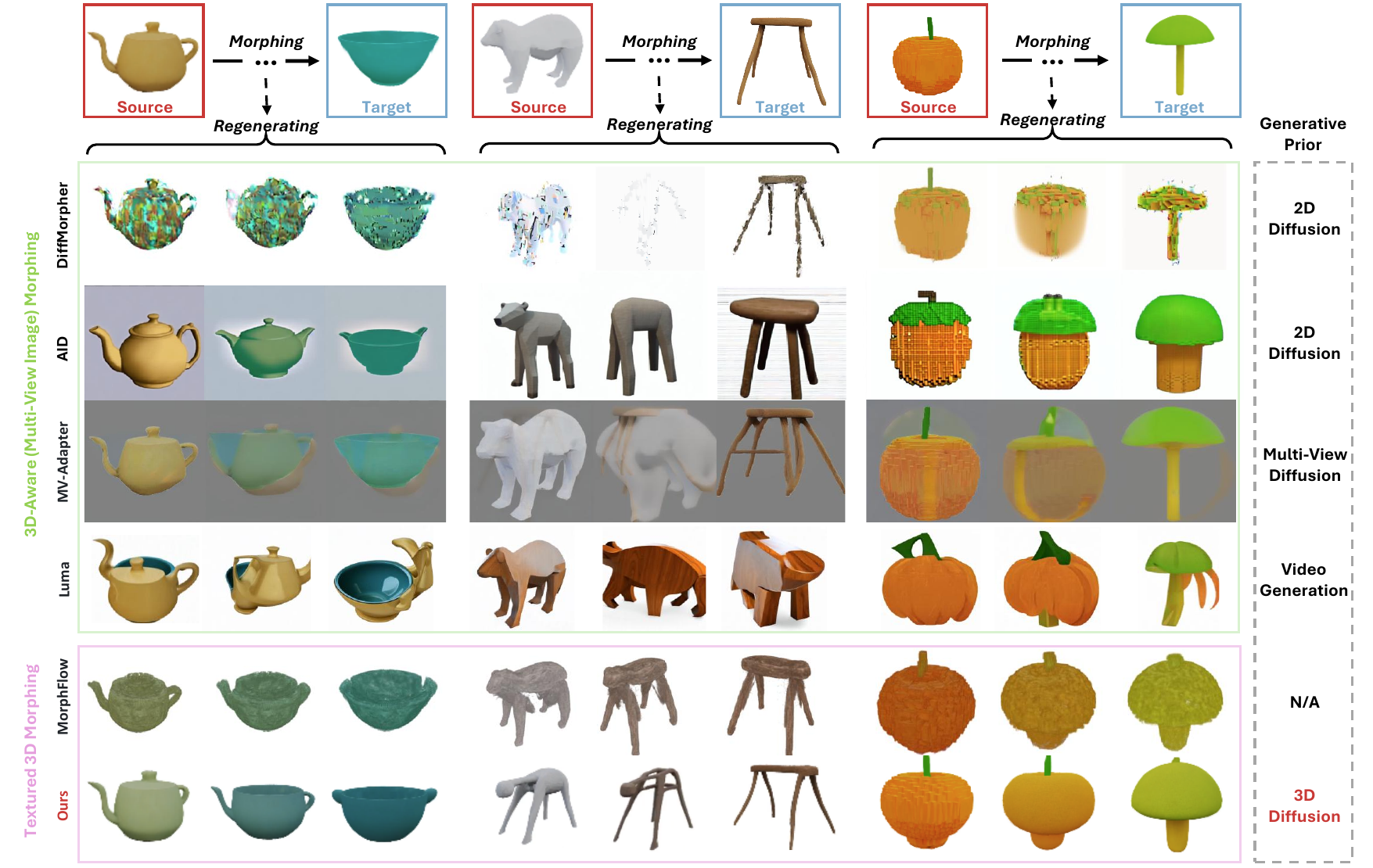}
    \vspace{-0.7cm}
    \caption{Qualitative comparisons of different methods from tasks, morphing tricks, and generative priors.  \textbf{\textit{More video results can be found
    \href{https://anonymous-888.github.io/siggraph25/}{\textcolor{red}{here}}.}}}
    \label{fig: baseline}
    \vspace{-0.2cm}
\end{figure*}

\subsubsection{Implementation}

To address this, as shown in Fig.~\ref{fig: framework}, we propose to enhance the low-frequency signal when generating the 3D interpolations. Specifically, the process is defined as:
\begin{align}
    F(h) &= \text{FFT}(h), \\
    F_{\omega<\omega_0}'(h) &= F_{\omega<\omega_0}(h) \odot scale, \\
    h' &= \text{IFFT}([F_{\omega<\omega_0}'(h),F_{\omega\geq\omega_0}(h)]),
\end{align}
where $h$ represents the tokens, $F(h)$ their Fourier features, and $h'$ the enhanced tokens, $\omega$ and $\omega_0$ denote Fourier frequencies and the threshold, with $\omega < \omega_0$ indicating low-frequency components. The $scale$ is the enhancement coefficient, and $\text{FFT}(\cdot)$ and $\text{IFFT}(\cdot)$ are the Fourier transform and inverse Fourier transform.

\section{Experiments}

\subsection{Implementation Details}

\subsubsection{3D Generation Model}

The 3D diffusion prior~\cite{lan2024gaussiananything} is trained on the G-Objaverse~\cite{deitke2023objaverse} dataset. Its geometry and texture diffusion models are based on the DiT architecture~\cite{chen2023pixart}, which consists of 24 layers, 16 attention heads, and a 1024-dimensional hidden space. The sparse point cloud $\mathbf{z}_G$ has a size of $M \times 3$ (with $M = 768$), and the corresponding feature $\mathbf{z}_T$ has dimensions $M \times 10$. All experiments use 250 denoising time steps. Fine-tuning is performed using LoRA models implemented with PEFT~\cite{peft} using 500 training steps (rank=16, alpha=20, and list=[`to\_k', `to\_q', `to\_v', `qkv']).

\subsubsection{Morphing Configuration} In our experiments, the initial value of $\alpha$ is sampled from a Beta distribution over the [0, 1] interval, with 10 points selected as interpolation weights. If multiple morphing is allowed, $\alpha$ can be concentrated in ranges with more variation by adjusting the Beta distribution or using the reschedule strategy from Diffmorpher. For Attention Fusion, we recommend starting from the first step, with geometry diffusion ending between steps 120 and 180, and texture diffusion finishing by step 5. For Token Reordering, it should begin at step 80 and end at step 200. For Low-Frequency Enhancement, the recommended time step range is from step 200 to step 230, while the $scale= 5$  and $\omega_0=0.1\pi$. \textit{More details can be found in the \textbf{Supplementary Materials}.}

\subsubsection{Baselines} \textit{(a) Task (Textured 3D Morphing)}: Morphflow~\cite{tsai2022multiview} directly implements morphing on 3D volumetric representations; \textit{(b) Morphing Tricks}: Diffmorpher~\cite{zhang2024diffmorpher} and AID~\cite{he2024aid} propose fusion strategies for attention to enhance smoothness; \textit{(c) Generative Priors}: We compare the performance of 2D diffusion (Diffmorpher, AID), multi-view diffusion (MV-Adapter~\cite{huang2024mvadapter}), video generation (Luma~\cite{luma2025dreammachine}), and 3D diffusion in the 3D morphing task.

\setlength{\tabcolsep}{2pt}
\begin{table}[t]
\caption{Quantitative comparisons of different methods.}
\vspace{-0.4cm}
\centering
\footnotesize
\begin{tabular}{cccccccc}
\toprule
\multirow{2}{*}{} & \multicolumn{5}{c}{Quantitative Metrics} & \multicolumn{2}{c}{User Study} \\ \cmidrule(lr){2-6} \cmidrule(lr){7-8}
 & FID$\downarrow$ &STP-GPT$\uparrow$ & SEP-GPT$\uparrow$ & PPL$\downarrow$ & PDV$\downarrow$ & STP-U$\uparrow$ & SEP-U$\uparrow$ \\ \midrule
DiffMorpher& 218.07 & 0.23 & 0.13 & 5.23  & 0.0535 & 0.435 &0.300\\
AID& 115.72 & 0.67 & 0.70 & 4.68 & 0.0118 &0.380 &0.505 \\
MV-Adapter& 120.93 & 0.63 & 0.57 & 7.29 & 0.0152  & 0.225 &0.350\\
Luma&  95.49 & 0.83 & 0.77 & 7.37 & 0.0007  & 0.415 &0.330\\ 
\rowcolor[HTML]{FFF2CC} 
MorphFlow& 147.70 & 0.87 & 0.90 & 3.10 & \textbf{0.0001}  & 0.555 &0.505\\
\rowcolor[HTML]{FFF2CC} 
Ours& \textbf{6.36} & \textbf{1.00} & \textbf{1.00} & \textbf{3.02} & \textbf{0.0001} & \textbf{0.915} &\textbf{0.950}\\
\bottomrule
\end{tabular}
\label{tab: evaluation}
\vspace{-0.3cm}
\end{table}

\subsubsection{Metrics}

The performance of textured 3D morphing is evaluated using the following metrics: \textit{(a) Fréchet Inception Distance (FID)}~\cite{heusel2017gans}: Fidelity is measured by comparing 1,000 images rendered from original and interpolated 3D representations. \textit{(b) Structural Plausibility-GPT (STP-GPT) and Semantic Plausibility-GPT (SEP-GPT)}: GPT-4o~\cite{gpt4} evaluates structural and semantic plausibility based on testing results, providing scores and explanations. \textit{(c) Perceptual Path Length (PPL) and Perceptual Distance Variance (PDV)}~\cite{zhang2024diffmorpher}: PPL sums perceptual losses over 20-frame sequences, reflecting smoothness and consistency, while PDV measures the variance, indicating transition homogeneity. \textit{(d) Structural Plausibility-User (STP-U) and Semantic Plausibility-User (SEP-U)}: Volunteers score morphing results based on structural and semantic plausibility.

\subsection{Evaluation}

\subsubsection{Textured 3D Morphing}

Our method adopts a 3D generation prior for 3D morphing, offering two main benefits: direct morphing of textured 3D representations, and maintaining structural and semantic consistency in interpolated representations, as shown in Fig.~\ref{fig:teaser} and Fig.~\ref{fig: ours_more}. The closest baseline to our method is MorphFlow~\cite{tsai2022multiview}, which morphs between two 3D volumetric representations using optimal transport optimization. However, as shown in Fig.\ref{fig: baseline} and Fig.~\ref{fig: baseline_1}, MorphFlow has two main drawbacks. First, the generated quality is low (See Tab.\ref{tab: evaluation}); the advantage of volumetric representations for photorealism is diminished due to the dense interpolation process, which requires morphing even points with no color. Second, it lacks an effective generative prior to the intermediate process, limiting its ability to understand the semantic meaning of intermediate stages. As a result, it often introduces artifacts in morphing scenarios, such as generating six legs instead of four when morphing a bear’s and a table’s legs.

Our method not only significantly outperforms MorphFlow in quality but also demonstrates an impressive ``understanding'' when interpolating across diverse 3D object pairs from different categories. This understanding is evident in two key aspects: (a) the effective mapping of semantically similar parts, as shown when morphing a boot into a red teddy bear, where our model smoothly splits the boot into legs and transforms them into the bear's facial features, and (b) minimal disconnected artifacts. The regenerative nature of our method ensures the fusion of source and target information while considering the distribution of the entire latent space, minimizing issues like 3D structure collapse or disconnected parts.

\subsubsection{3D-Aware (Multi-View Image) Morphing}

We evaluated alternatives to our method for textured 3D morphing from two perspectives. First, 3D morphing can be viewed as multi-view image morphing, where multi-view source and target images are fed into image morphing methods for pseudo-3D morphing. Second, many image and video generation models, beyond 3D priors, can perform interpolation tasks. Notably, 2D generation models~\cite{yang2023context,yang2024learning,yang2023designing,shi2023mvdream,huang2024mvadapter}, trained on larger datasets often produce 3D-consistent images with superior semantic, structural, and textural understanding compared to 3D models. Based on these insights, we explored three types of generative priors: 2D diffusion, multi-view diffusion, and video generation.

As shown in Fig.~\ref{fig: baseline} and Tab.~\ref{tab: evaluation}, compared to state-of-the-art 2D image morphing methods, such as DiffMorpher~\cite{zhang2024diffmorpher} and AID~\cite{he2024aid}, we found that 2D diffusion models often suffer from mode collapse when influenced by non-object image regions (e.g., DiffMorpher is sensitive to white backgrounds). Their lack of 3D consistency leads to inconsistent morphing results for the same $\alpha$ across different viewpoints. When comparing with multi-view diffusion models, we observed that their image-based multi-view generation is limited by pixel-aligned morphing. This limitation becomes apparent when matching pixels across large spatial distances—such as aligning a point on the lower edge of a pumpkin's surface with a point on the stem of a mushroom in the image's center, resulting in interpolation errors. Lastly, video generation models, while demonstrating strong spatial understanding and achieving morphing by specifying the source and target images as the first and last frames, suffer from limited controllability during generation, often producing incomplete or out-of-frame content. Additionally, the structural consistency of intermediate frames remains suboptimal.

\begin{figure}[t]
    \centering
    \includegraphics[width=0.49\textwidth]{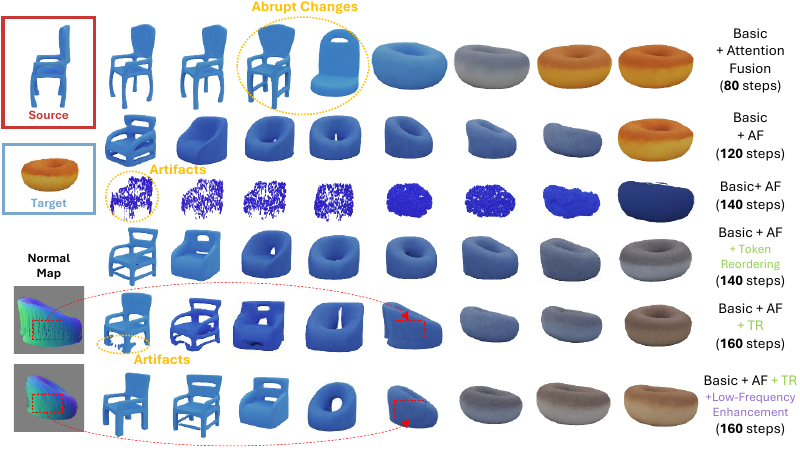}
   \vspace{-0.6cm}
    \caption{Ablation experiments of our proposed strategies.}
    \label{fig: ablation}
    \vspace{-0.6cm}
\end{figure}

\subsection{Ablation Study}

Balancing smooth transitions with structural plausibility (or overall generation quality) is challenging, especially in selecting the time step range for Attention Fusion. We empirically observe that texture diffusion allows for minimal attention fusion, while geometry diffusion offers a larger operational range, aligning well with the importance of shape understanding in 3D morphing. To address this, we incorporate Attention Fusion in texture diffusion from the first to the fifth time step, while progressively extending the final time step for geometry diffusion. Adding Attention Fusion on top of basic interpolation improves smoothness, but extending the final time step too far causes 3D structural collapse. Token Reordering helps mitigate this issue, though pushing the time step further reduces generative quality. Ultimately, by applying the Low-Frequency Enhancement strategy, we balance the smoothness and structural plausibility and ensure all frequencies are maintained effectively. More details can be found in  Fig.~\ref{fig: ablation} and \textbf{\textit{Supplementary Materials}}.

\section{Conclusions}

 We propose a method to achieve smooth and plausible morphing sequences across diverse cross-category 3D object pairs, incorporating Attention Fusion, Token Reordering, and Low-Frequency Enhancement. This introduces a new paradigm for textured 3D morphing, extending beyond the limitations of previous research  confined to shape-only morphing on topologically aligned datasets.

\noindent \textit{\textbf{Discussions.}} Our future work will focus on morphing more complex textured 3D objects, exploring two main directions: (a) Enhancing fidelity and diversity using advanced 3D generation models like Trellis~\cite{xiang2024structured}, and (b) Expanding morphing to generate complex sequences, such as few-shot motion interpolation~\cite{shen2024dreammover}, while maintaining temporal consistency. Additionally, we aim to explore the transitions for 4D content like the "Birth and Death of a Rose"~\cite{geng2024birthdeathrose}.


\begin{figure*}[p]
    \centering
    \includegraphics[width=\textwidth]{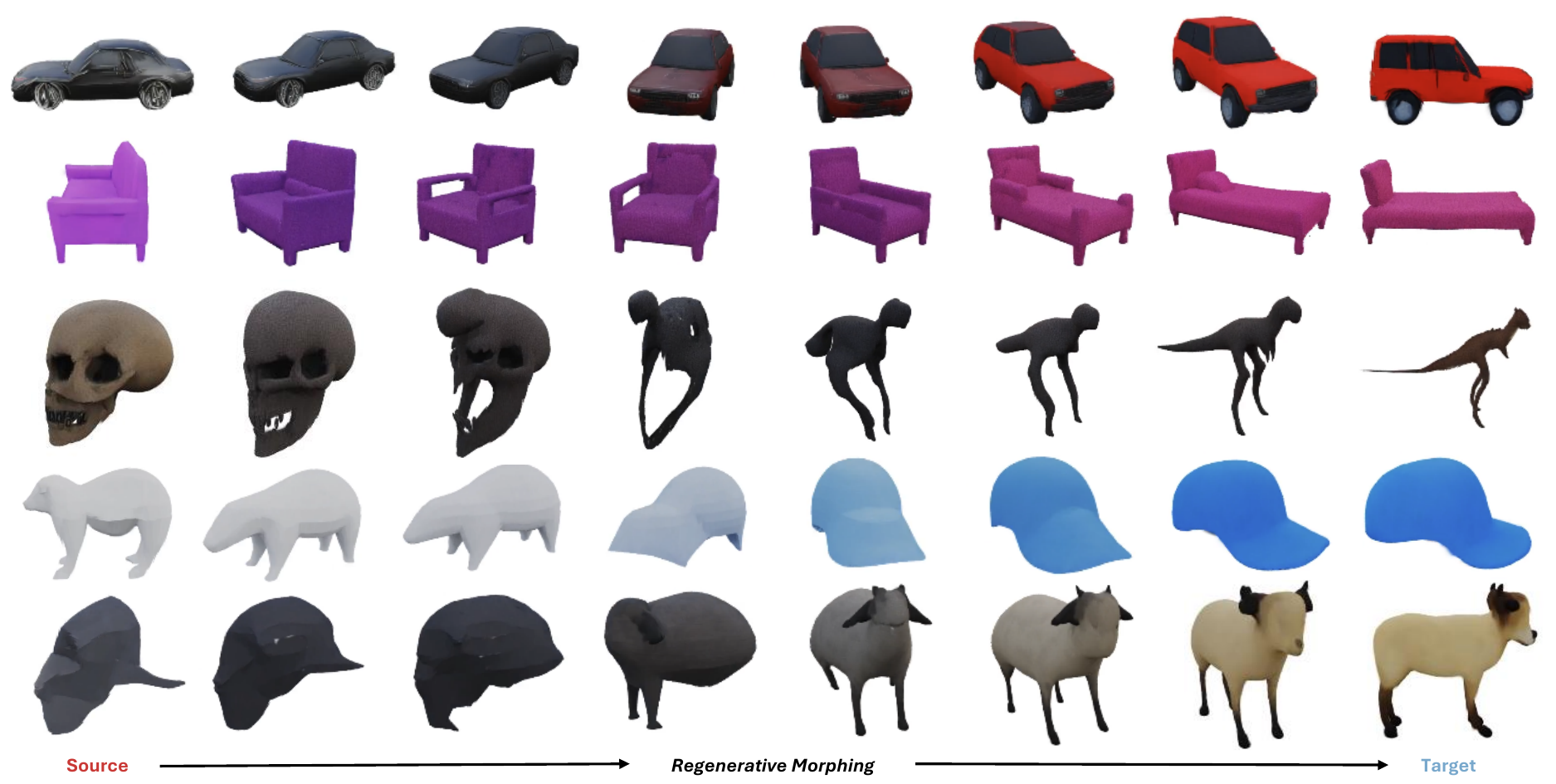}
    \vspace{-0.6cm}
    \caption{More 3D morphing sequences generated by our method.}
    \label{fig: ours_more}
 
\end{figure*}

\begin{figure*}[p]
    \centering
    \includegraphics[width=1.0\textwidth]{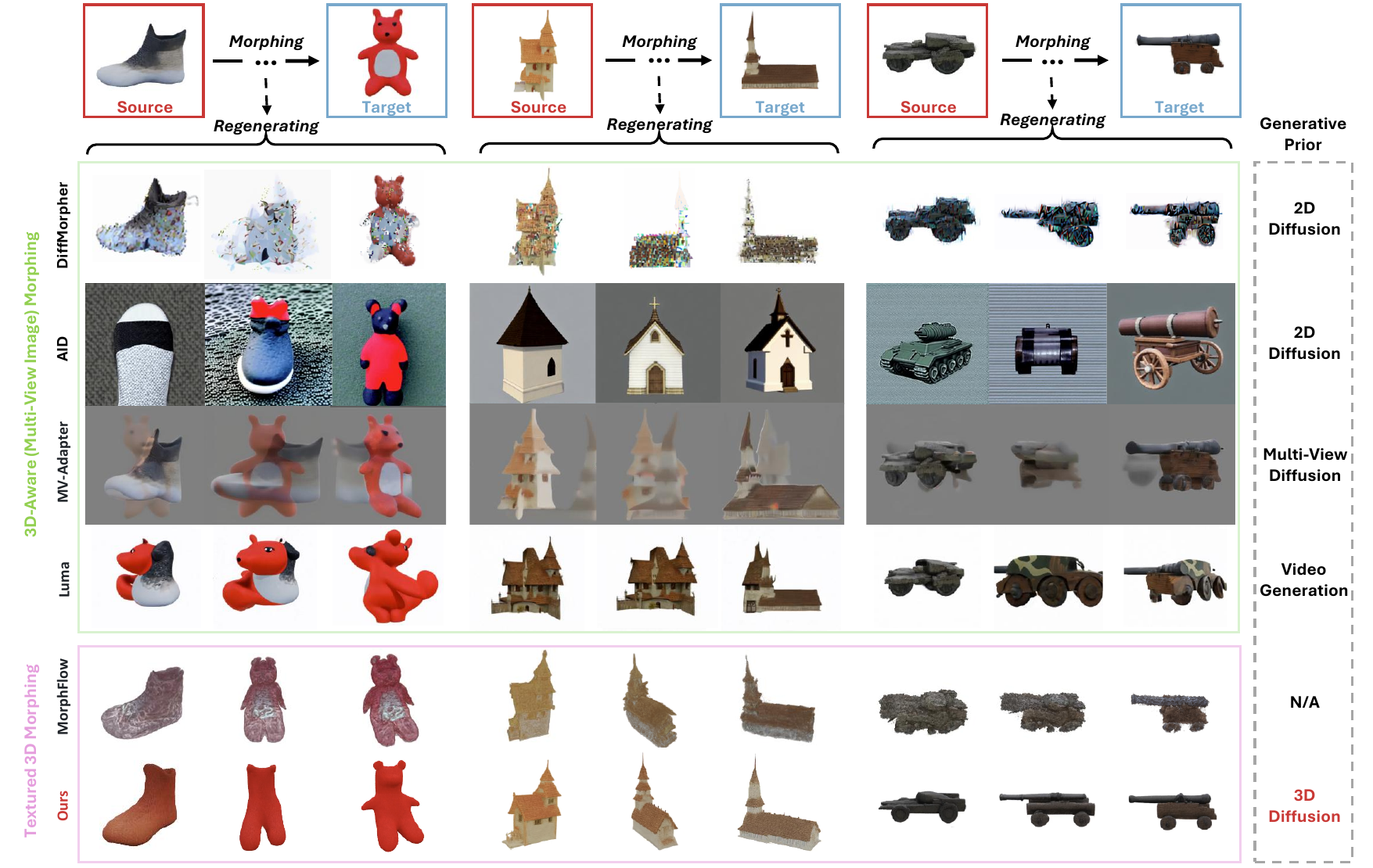}
    \vspace{-0.6cm}
    \caption{More qualitative comparisons of different methods from tasks, morphing tricks, and generative priors.  \textbf{\textit{More video results can be found
    \href{https://songlin1998.github.io/Textured-3D-Morphing/}{\textcolor{red}{here}}.}}}
    \label{fig: baseline_1}
    \vspace{-0.2cm}
\end{figure*}

\bibliographystyle{ACM-Reference-Format}
\bibliography{sample-bibliography}


\begin{thebibliography}{84}


\ifx \showCODEN    \undefined \def \showCODEN     #1{\unskip}     \fi
\ifx \showDOI      \undefined \def \showDOI       #1{#1}\fi
\ifx \showISBNx    \undefined \def \showISBNx     #1{\unskip}     \fi
\ifx \showISBNxiii \undefined \def \showISBNxiii  #1{\unskip}     \fi
\ifx \showISSN     \undefined \def \showISSN      #1{\unskip}     \fi
\ifx \showLCCN     \undefined \def \showLCCN      #1{\unskip}     \fi
\ifx \shownote     \undefined \def \shownote      #1{#1}          \fi
\ifx \showarticletitle \undefined \def \showarticletitle #1{#1}   \fi
\ifx \showURL      \undefined \def \showURL       {\relax}        \fi
\providecommand\bibfield[2]{#2}
\providecommand\bibinfo[2]{#2}
\providecommand\natexlab[1]{#1}
\providecommand\showeprint[2][]{arXiv:#2}

\bibitem[\protect\citeauthoryear{Abdelreheem, Skorokhodov, Ovsjanikov, and
  Wonka}{Abdelreheem et~al\mbox{.}}{2023}]%
        {abdelreheem2023satr}
\bibfield{author}{\bibinfo{person}{Ahmed Abdelreheem}, \bibinfo{person}{Ivan
  Skorokhodov}, \bibinfo{person}{Maks Ovsjanikov}, {and} \bibinfo{person}{Peter
  Wonka}.} \bibinfo{year}{2023}\natexlab{}.
\newblock \showarticletitle{Satr: Zero-shot semantic segmentation of 3d
  shapes}. In \bibinfo{booktitle}{\emph{Proceedings of the IEEE/CVF
  International Conference on Computer Vision}}. \bibinfo{pages}{15166--15179}.
\newblock


\bibitem[\protect\citeauthoryear{AI}{AI}{2025}]%
        {luma2025dreammachine}
\bibfield{author}{\bibinfo{person}{Luma~Labs AI}.}
  \bibinfo{year}{2025}\natexlab{}.
\newblock \bibinfo{title}{Luma Dream Machine: AI-Powered Video Content
  Creation}.
\newblock \bibinfo{howpublished}{\url{https://dream-machine.lumalabs.ai/}}.
\newblock
\newblock
\shownote{Accessed: 2025-01-15.}


\bibitem[\protect\citeauthoryear{Albergo, Boffi, and Vanden-Eijnden}{Albergo
  et~al\mbox{.}}{2023}]%
        {albergo2023stochastic}
\bibfield{author}{\bibinfo{person}{Michael~S Albergo},
  \bibinfo{person}{Nicholas~M Boffi}, {and} \bibinfo{person}{Eric
  Vanden-Eijnden}.} \bibinfo{year}{2023}\natexlab{}.
\newblock \showarticletitle{Stochastic interpolants: A unifying framework for
  flows and diffusions}.
\newblock \bibinfo{journal}{\emph{arXiv preprint arXiv:2303.08797}}
  (\bibinfo{year}{2023}).
\newblock


\bibitem[\protect\citeauthoryear{Aloraibi}{Aloraibi}{2023}]%
        {aloraibi2023image}
\bibfield{author}{\bibinfo{person}{Alyaa~Qusay Aloraibi}.}
  \bibinfo{year}{2023}\natexlab{}.
\newblock \showarticletitle{Image morphing techniques: A review}.
\newblock  (\bibinfo{year}{2023}).
\newblock


\bibitem[\protect\citeauthoryear{Averbuch-Elor, Cohen-Or, and
  Kopf}{Averbuch-Elor et~al\mbox{.}}{2016}]%
        {averbuch2016smooth}
\bibfield{author}{\bibinfo{person}{Hadar Averbuch-Elor},
  \bibinfo{person}{Daniel Cohen-Or}, {and} \bibinfo{person}{Johannes Kopf}.}
  \bibinfo{year}{2016}\natexlab{}.
\newblock \showarticletitle{Smooth image sequences for data-driven morphing}.
  In \bibinfo{booktitle}{\emph{computer graphics forum}},
  Vol.~\bibinfo{volume}{35}. Wiley Online Library, \bibinfo{pages}{203--213}.
\newblock


\bibitem[\protect\citeauthoryear{Ayd{\i}nl{\i}lar and
  Sahillio{\u{g}}lu}{Ayd{\i}nl{\i}lar and Sahillio{\u{g}}lu}{2021}]%
        {aydinlilar2021part}
\bibfield{author}{\bibinfo{person}{Melike Ayd{\i}nl{\i}lar} {and}
  \bibinfo{person}{Yusuf Sahillio{\u{g}}lu}.} \bibinfo{year}{2021}\natexlab{}.
\newblock \showarticletitle{Part-based data-driven 3D shape interpolation}.
\newblock \bibinfo{journal}{\emph{Computer-Aided Design}}
  \bibinfo{volume}{136} (\bibinfo{year}{2021}), \bibinfo{pages}{103027}.
\newblock


\bibitem[\protect\citeauthoryear{Beier and Neely}{Beier and Neely}{2023}]%
        {beier2023feature}
\bibfield{author}{\bibinfo{person}{Thaddeus Beier} {and} \bibinfo{person}{Shawn
  Neely}.} \bibinfo{year}{2023}\natexlab{}.
\newblock \showarticletitle{Feature-based image metamorphosis}.
\newblock In \bibinfo{booktitle}{\emph{Seminal Graphics Papers: Pushing the
  Boundaries, Volume 2}}. \bibinfo{pages}{529--536}.
\newblock


\bibitem[\protect\citeauthoryear{Bhatt}{Bhatt}{2011}]%
        {bhatt2011comparative}
\bibfield{author}{\bibinfo{person}{Bhumika~G Bhatt}.}
  \bibinfo{year}{2011}\natexlab{}.
\newblock \showarticletitle{Comparative study of triangulation based and
  feature based image morphing}.
\newblock \bibinfo{journal}{\emph{Signal \& Image Processing}}
  \bibinfo{volume}{2}, \bibinfo{number}{4} (\bibinfo{year}{2011}),
  \bibinfo{pages}{235}.
\newblock


\bibitem[\protect\citeauthoryear{Bogo, Romero, Loper, and Black}{Bogo
  et~al\mbox{.}}{2014}]%
        {bogo2014faust}
\bibfield{author}{\bibinfo{person}{Federica Bogo}, \bibinfo{person}{Javier
  Romero}, \bibinfo{person}{Matthew Loper}, {and} \bibinfo{person}{Michael~J
  Black}.} \bibinfo{year}{2014}\natexlab{}.
\newblock \showarticletitle{FAUST: Dataset and evaluation for 3D mesh
  registration}. In \bibinfo{booktitle}{\emph{Proceedings of the IEEE
  conference on computer vision and pattern recognition}}.
  \bibinfo{pages}{3794--3801}.
\newblock


\bibitem[\protect\citeauthoryear{Chen, Yu, Ge, Yao, Xie, Wu, Wang, Kwok, Luo,
  Lu, et~al\mbox{.}}{Chen et~al\mbox{.}}{2024b}]%
        {chen2023pixart}
\bibfield{author}{\bibinfo{person}{Junsong Chen}, \bibinfo{person}{Jincheng
  Yu}, \bibinfo{person}{Chongjian Ge}, \bibinfo{person}{Lewei Yao},
  \bibinfo{person}{Enze Xie}, \bibinfo{person}{Yue Wu},
  \bibinfo{person}{Zhongdao Wang}, \bibinfo{person}{James Kwok},
  \bibinfo{person}{Ping Luo}, \bibinfo{person}{Huchuan Lu}, {et~al\mbox{.}}}
  \bibinfo{year}{2024}\natexlab{b}.
\newblock \showarticletitle{Pixart-$\alpha$: Fast training of diffusion
  transformer for photorealistic text-to-image synthesis}.
\newblock \bibinfo{journal}{\emph{ICLR}} (\bibinfo{year}{2024}).
\newblock


\bibitem[\protect\citeauthoryear{Chen, Tang, Dong, Cao, Hong, Lan, Wang, Xie,
  Wu, Saito, et~al\mbox{.}}{Chen et~al\mbox{.}}{2024a}]%
        {chen20243dtopia}
\bibfield{author}{\bibinfo{person}{Zhaoxi Chen}, \bibinfo{person}{Jiaxiang
  Tang}, \bibinfo{person}{Yuhao Dong}, \bibinfo{person}{Ziang Cao},
  \bibinfo{person}{Fangzhou Hong}, \bibinfo{person}{Yushi Lan},
  \bibinfo{person}{Tengfei Wang}, \bibinfo{person}{Haozhe Xie},
  \bibinfo{person}{Tong Wu}, \bibinfo{person}{Shunsuke Saito}, {et~al\mbox{.}}}
  \bibinfo{year}{2024}\natexlab{a}.
\newblock \showarticletitle{3dtopia-xl: Scaling high-quality 3d asset
  generation via primitive diffusion}.
\newblock \bibinfo{journal}{\emph{arXiv preprint arXiv:2409.12957}}
  (\bibinfo{year}{2024}).
\newblock


\bibitem[\protect\citeauthoryear{Darabi, Shechtman, Barnes, Goldman, and
  Sen}{Darabi et~al\mbox{.}}{2012}]%
        {darabi2012image}
\bibfield{author}{\bibinfo{person}{Soheil Darabi}, \bibinfo{person}{Eli
  Shechtman}, \bibinfo{person}{Connelly Barnes}, \bibinfo{person}{Dan~B
  Goldman}, {and} \bibinfo{person}{Pradeep Sen}.}
  \bibinfo{year}{2012}\natexlab{}.
\newblock \showarticletitle{Image melding: Combining inconsistent images using
  patch-based synthesis}.
\newblock \bibinfo{journal}{\emph{ACM Transactions on graphics (TOG)}}
  \bibinfo{volume}{31}, \bibinfo{number}{4} (\bibinfo{year}{2012}),
  \bibinfo{pages}{1--10}.
\newblock


\bibitem[\protect\citeauthoryear{Deitke, Schwenk, Salvador, Weihs, Michel,
  VanderBilt, Schmidt, Ehsani, Kembhavi, and Farhadi}{Deitke
  et~al\mbox{.}}{2023}]%
        {deitke2023objaverse}
\bibfield{author}{\bibinfo{person}{Matt Deitke}, \bibinfo{person}{Dustin
  Schwenk}, \bibinfo{person}{Jordi Salvador}, \bibinfo{person}{Luca Weihs},
  \bibinfo{person}{Oscar Michel}, \bibinfo{person}{Eli VanderBilt},
  \bibinfo{person}{Ludwig Schmidt}, \bibinfo{person}{Kiana Ehsani},
  \bibinfo{person}{Aniruddha Kembhavi}, {and} \bibinfo{person}{Ali Farhadi}.}
  \bibinfo{year}{2023}\natexlab{}.
\newblock \showarticletitle{Objaverse: A universe of annotated 3d objects}. In
  \bibinfo{booktitle}{\emph{Proceedings of the IEEE/CVF Conference on Computer
  Vision and Pattern Recognition}}. \bibinfo{pages}{13142--13153}.
\newblock


\bibitem[\protect\citeauthoryear{Deng, Wang, Lu, He, Zhang, Yu, and Zhang}{Deng
  et~al\mbox{.}}{2023}]%
        {deng2023se}
\bibfield{author}{\bibinfo{person}{Jiacheng Deng}, \bibinfo{person}{Chuxin
  Wang}, \bibinfo{person}{Jiahao Lu}, \bibinfo{person}{Jianfeng He},
  \bibinfo{person}{Tianzhu Zhang}, \bibinfo{person}{Jiyang Yu}, {and}
  \bibinfo{person}{Zhe Zhang}.} \bibinfo{year}{2023}\natexlab{}.
\newblock \showarticletitle{Se-ornet: Self-ensembling orientation-aware network
  for unsupervised point cloud shape correspondence}. In
  \bibinfo{booktitle}{\emph{Proceedings of the IEEE/CVF Conference on Computer
  Vision and Pattern Recognition}}. \bibinfo{pages}{5364--5373}.
\newblock


\bibitem[\protect\citeauthoryear{Dyke, Lai, Rosin, Zappal{\`a}, Dykes, Guo, Li,
  Marin, Melzi, and Yang}{Dyke et~al\mbox{.}}{2020}]%
        {dyke2020shrec}
\bibfield{author}{\bibinfo{person}{Roberto~M Dyke}, \bibinfo{person}{Yu-Kun
  Lai}, \bibinfo{person}{Paul~L Rosin}, \bibinfo{person}{Stefano Zappal{\`a}},
  \bibinfo{person}{Seana Dykes}, \bibinfo{person}{Daoliang Guo},
  \bibinfo{person}{Kun Li}, \bibinfo{person}{Riccardo Marin},
  \bibinfo{person}{Simone Melzi}, {and} \bibinfo{person}{Jingyu Yang}.}
  \bibinfo{year}{2020}\natexlab{}.
\newblock \showarticletitle{SHREC’20: Shape correspondence with non-isometric
  deformations}.
\newblock \bibinfo{journal}{\emph{Computers \& Graphics}}  \bibinfo{volume}{92}
  (\bibinfo{year}{2020}), \bibinfo{pages}{28--43}.
\newblock


\bibitem[\protect\citeauthoryear{Edelstein, Ezuz, and Ben-Chen}{Edelstein
  et~al\mbox{.}}{2019}]%
        {edelstein2019enigma}
\bibfield{author}{\bibinfo{person}{Michal Edelstein}, \bibinfo{person}{Danielle
  Ezuz}, {and} \bibinfo{person}{Mirela Ben-Chen}.}
  \bibinfo{year}{2019}\natexlab{}.
\newblock \showarticletitle{Enigma: Evolutionary non-isometric geometry
  matching}.
\newblock \bibinfo{journal}{\emph{arXiv preprint arXiv:1905.10763}}
  (\bibinfo{year}{2019}).
\newblock


\bibitem[\protect\citeauthoryear{Eisenberger, Lahner, and Cremers}{Eisenberger
  et~al\mbox{.}}{2020}]%
        {eisenberger2020smooth}
\bibfield{author}{\bibinfo{person}{Marvin Eisenberger}, \bibinfo{person}{Zorah
  Lahner}, {and} \bibinfo{person}{Daniel Cremers}.}
  \bibinfo{year}{2020}\natexlab{}.
\newblock \showarticletitle{Smooth shells: Multi-scale shape registration with
  functional maps}. In \bibinfo{booktitle}{\emph{Proceedings of the IEEE/CVF
  Conference on Computer Vision and Pattern Recognition}}.
  \bibinfo{pages}{12265--12274}.
\newblock


\bibitem[\protect\citeauthoryear{Eisenberger, Novotny, Kerchenbaum, Labatut,
  Neverova, Cremers, and Vedaldi}{Eisenberger et~al\mbox{.}}{2021}]%
        {eisenberger2021neuromorph}
\bibfield{author}{\bibinfo{person}{Marvin Eisenberger}, \bibinfo{person}{David
  Novotny}, \bibinfo{person}{Gael Kerchenbaum}, \bibinfo{person}{Patrick
  Labatut}, \bibinfo{person}{Natalia Neverova}, \bibinfo{person}{Daniel
  Cremers}, {and} \bibinfo{person}{Andrea Vedaldi}.}
  \bibinfo{year}{2021}\natexlab{}.
\newblock \showarticletitle{Neuromorph: Unsupervised shape interpolation and
  correspondence in one go}. In \bibinfo{booktitle}{\emph{Proceedings of the
  IEEE/CVF Conference on Computer Vision and Pattern Recognition}}.
  \bibinfo{pages}{7473--7483}.
\newblock


\bibitem[\protect\citeauthoryear{Fish, Zhang, Perry, Cohen-Or, Shechtman, and
  Barnes}{Fish et~al\mbox{.}}{2020}]%
        {fish2020image}
\bibfield{author}{\bibinfo{person}{Noa Fish}, \bibinfo{person}{Richard Zhang},
  \bibinfo{person}{Lilach Perry}, \bibinfo{person}{Daniel Cohen-Or},
  \bibinfo{person}{Eli Shechtman}, {and} \bibinfo{person}{Connelly Barnes}.}
  \bibinfo{year}{2020}\natexlab{}.
\newblock \showarticletitle{Image morphing with perceptual constraints and stn
  alignment}. In \bibinfo{booktitle}{\emph{Computer Graphics Forum}},
  Vol.~\bibinfo{volume}{39}. Wiley Online Library, \bibinfo{pages}{303--313}.
\newblock


\bibitem[\protect\citeauthoryear{Fulton, Modi, Duvenaud, Levin, and
  Jacobson}{Fulton et~al\mbox{.}}{2019}]%
        {fulton2019latent}
\bibfield{author}{\bibinfo{person}{Lawson Fulton}, \bibinfo{person}{Vismay
  Modi}, \bibinfo{person}{David Duvenaud}, \bibinfo{person}{David~IW Levin},
  {and} \bibinfo{person}{Alec Jacobson}.} \bibinfo{year}{2019}\natexlab{}.
\newblock \showarticletitle{Latent-space dynamics for reduced deformable
  simulation}. In \bibinfo{booktitle}{\emph{Computer graphics forum}},
  Vol.~\bibinfo{volume}{38}. Wiley Online Library, \bibinfo{pages}{379--391}.
\newblock


\bibitem[\protect\citeauthoryear{Gao, Aigerman, Groueix, Kim, and Hanocka}{Gao
  et~al\mbox{.}}{2023}]%
        {gao2023textdeformer}
\bibfield{author}{\bibinfo{person}{William Gao}, \bibinfo{person}{Noam
  Aigerman}, \bibinfo{person}{Thibault Groueix}, \bibinfo{person}{Vova Kim},
  {and} \bibinfo{person}{Rana Hanocka}.} \bibinfo{year}{2023}\natexlab{}.
\newblock \showarticletitle{Textdeformer: Geometry manipulation using text
  guidance}. In \bibinfo{booktitle}{\emph{ACM SIGGRAPH 2023 Conference
  Proceedings}}. \bibinfo{pages}{1--11}.
\newblock


\bibitem[\protect\citeauthoryear{Geng, Zhang, Wu, and Wu}{Geng
  et~al\mbox{.}}{2024}]%
        {geng2024birthdeathrose}
\bibfield{author}{\bibinfo{person}{Chen Geng}, \bibinfo{person}{Yunzhi Zhang},
  \bibinfo{person}{Shangzhe Wu}, {and} \bibinfo{person}{Jiajun Wu}.}
  \bibinfo{year}{2024}\natexlab{}.
\newblock \bibinfo{title}{Birth and Death of a Rose}.
\newblock
\newblock
\showeprint[arxiv]{2412.05278}~[cs.CV]
\urldef\tempurl%
\url{https://arxiv.org/abs/2412.05278}
\showURL{%
\tempurl}


\bibitem[\protect\citeauthoryear{Haque, Tancik, Efros, Holynski, and
  Kanazawa}{Haque et~al\mbox{.}}{2023}]%
        {haque2023instruct}
\bibfield{author}{\bibinfo{person}{Ayaan Haque}, \bibinfo{person}{Matthew
  Tancik}, \bibinfo{person}{Alexei~A Efros}, \bibinfo{person}{Aleksander
  Holynski}, {and} \bibinfo{person}{Angjoo Kanazawa}.}
  \bibinfo{year}{2023}\natexlab{}.
\newblock \showarticletitle{Instruct-nerf2nerf: Editing 3d scenes with
  instructions}. In \bibinfo{booktitle}{\emph{Proceedings of the IEEE/CVF
  International Conference on Computer Vision}}. \bibinfo{pages}{19740--19750}.
\newblock


\bibitem[\protect\citeauthoryear{He, Wang, Liu, and Yao}{He
  et~al\mbox{.}}{2024}]%
        {he2024aid}
\bibfield{author}{\bibinfo{person}{Qiyuan He}, \bibinfo{person}{Jinghao Wang},
  \bibinfo{person}{Ziwei Liu}, {and} \bibinfo{person}{Angela Yao}.}
  \bibinfo{year}{2024}\natexlab{}.
\newblock \showarticletitle{AID: Attention Interpolation of Text-to-Image
  Diffusion}.
\newblock \bibinfo{journal}{\emph{NeurIPS}} (\bibinfo{year}{2024}).
\newblock


\bibitem[\protect\citeauthoryear{Hedlin, Sharma, Mahajan, Isack, Kar,
  Tagliasacchi, and Yi}{Hedlin et~al\mbox{.}}{2024}]%
        {hedlin2024unsupervised}
\bibfield{author}{\bibinfo{person}{Eric Hedlin}, \bibinfo{person}{Gopal
  Sharma}, \bibinfo{person}{Shweta Mahajan}, \bibinfo{person}{Hossam Isack},
  \bibinfo{person}{Abhishek Kar}, \bibinfo{person}{Andrea Tagliasacchi}, {and}
  \bibinfo{person}{Kwang~Moo Yi}.} \bibinfo{year}{2024}\natexlab{}.
\newblock \showarticletitle{Unsupervised semantic correspondence using stable
  diffusion}.
\newblock \bibinfo{journal}{\emph{Advances in Neural Information Processing
  Systems}}  \bibinfo{volume}{36} (\bibinfo{year}{2024}).
\newblock


\bibitem[\protect\citeauthoryear{Heusel, Ramsauer, Unterthiner, Nessler, and
  Hochreiter}{Heusel et~al\mbox{.}}{2017}]%
        {heusel2017gans}
\bibfield{author}{\bibinfo{person}{Martin Heusel}, \bibinfo{person}{Hubert
  Ramsauer}, \bibinfo{person}{Thomas Unterthiner}, \bibinfo{person}{Bernhard
  Nessler}, {and} \bibinfo{person}{Sepp Hochreiter}.}
  \bibinfo{year}{2017}\natexlab{}.
\newblock \showarticletitle{Gans trained by a two time-scale update rule
  converge to a local nash equilibrium}.
\newblock \bibinfo{journal}{\emph{Advances in neural information processing
  systems}}  \bibinfo{volume}{30} (\bibinfo{year}{2017}).
\newblock


\bibitem[\protect\citeauthoryear{Hu, Shen, Wallis, Allen-Zhu, Li, Wang, Wang,
  and Chen}{Hu et~al\mbox{.}}{2021}]%
        {hu2021lora}
\bibfield{author}{\bibinfo{person}{Edward~J Hu}, \bibinfo{person}{Yelong Shen},
  \bibinfo{person}{Phillip Wallis}, \bibinfo{person}{Zeyuan Allen-Zhu},
  \bibinfo{person}{Yuanzhi Li}, \bibinfo{person}{Shean Wang},
  \bibinfo{person}{Lu Wang}, {and} \bibinfo{person}{Weizhu Chen}.}
  \bibinfo{year}{2021}\natexlab{}.
\newblock \showarticletitle{Lora: Low-rank adaptation of large language
  models}.
\newblock \bibinfo{journal}{\emph{arXiv preprint arXiv:2106.09685}}
  (\bibinfo{year}{2021}).
\newblock


\bibitem[\protect\citeauthoryear{Huang, Yu, Chen, Geiger, and Gao}{Huang
  et~al\mbox{.}}{2024c}]%
        {huang20242d}
\bibfield{author}{\bibinfo{person}{Binbin Huang}, \bibinfo{person}{Zehao Yu},
  \bibinfo{person}{Anpei Chen}, \bibinfo{person}{Andreas Geiger}, {and}
  \bibinfo{person}{Shenghua Gao}.} \bibinfo{year}{2024}\natexlab{c}.
\newblock \showarticletitle{2d gaussian splatting for geometrically accurate
  radiance fields}. In \bibinfo{booktitle}{\emph{ACM SIGGRAPH 2024 conference
  papers}}. \bibinfo{pages}{1--11}.
\newblock


\bibitem[\protect\citeauthoryear{Huang, Guo, Wang, Yi, Ma, Cao, and
  Sheng}{Huang et~al\mbox{.}}{2024a}]%
        {huang2024mvadapter}
\bibfield{author}{\bibinfo{person}{Zehuan Huang}, \bibinfo{person}{Yuanchen
  Guo}, \bibinfo{person}{Haoran Wang}, \bibinfo{person}{Ran Yi},
  \bibinfo{person}{Lizhuang Ma}, \bibinfo{person}{Yan-Pei Cao}, {and}
  \bibinfo{person}{Lu Sheng}.} \bibinfo{year}{2024}\natexlab{a}.
\newblock \showarticletitle{MV-Adapter: Multi-view Consistent Image Generation
  Made Easy}.
\newblock \bibinfo{journal}{\emph{arXiv preprint arXiv:2412.03632}}
  (\bibinfo{year}{2024}).
\newblock


\bibitem[\protect\citeauthoryear{Huang, Johnson, Debnath, Rehg, and Wu}{Huang
  et~al\mbox{.}}{2024b}]%
        {huang2024pointinfinity}
\bibfield{author}{\bibinfo{person}{Zixuan Huang}, \bibinfo{person}{Justin
  Johnson}, \bibinfo{person}{Shoubhik Debnath}, \bibinfo{person}{James~M Rehg},
  {and} \bibinfo{person}{Chao-Yuan Wu}.} \bibinfo{year}{2024}\natexlab{b}.
\newblock \showarticletitle{PointInfinity: Resolution-Invariant Point Diffusion
  Models}. In \bibinfo{booktitle}{\emph{Proceedings of the IEEE/CVF Conference
  on Computer Vision and Pattern Recognition}}. \bibinfo{pages}{10050--10060}.
\newblock


\bibitem[\protect\citeauthoryear{Jacobson, Deng, Kavan, and Lewis}{Jacobson
  et~al\mbox{.}}{2014}]%
        {jacobson2014skinning}
\bibfield{author}{\bibinfo{person}{Alec Jacobson}, \bibinfo{person}{Zhigang
  Deng}, \bibinfo{person}{Ladislav Kavan}, {and} \bibinfo{person}{John~P
  Lewis}.} \bibinfo{year}{2014}\natexlab{}.
\newblock \showarticletitle{Skinning: Real-time shape deformation (full text
  not available)}.
\newblock In \bibinfo{booktitle}{\emph{ACM SIGGRAPH 2014 Courses}}.
  \bibinfo{pages}{1--1}.
\newblock


\bibitem[\protect\citeauthoryear{Kim, Lang, Aigerman, Groueix, Kim, and
  Hanocka}{Kim et~al\mbox{.}}{2024}]%
        {kim2024meshup}
\bibfield{author}{\bibinfo{person}{Hyunwoo Kim}, \bibinfo{person}{Itai Lang},
  \bibinfo{person}{Noam Aigerman}, \bibinfo{person}{Thibault Groueix},
  \bibinfo{person}{Vladimir~G Kim}, {and} \bibinfo{person}{Rana Hanocka}.}
  \bibinfo{year}{2024}\natexlab{}.
\newblock \showarticletitle{MeshUp: Multi-Target Mesh Deformation via Blended
  Score Distillation}.
\newblock \bibinfo{journal}{\emph{arXiv preprint arXiv:2408.14899}}
  (\bibinfo{year}{2024}).
\newblock


\bibitem[\protect\citeauthoryear{Kim, Lipman, and Funkhouser}{Kim
  et~al\mbox{.}}{2011}]%
        {kim2011blended}
\bibfield{author}{\bibinfo{person}{Vladimir~G Kim}, \bibinfo{person}{Yaron
  Lipman}, {and} \bibinfo{person}{Thomas Funkhouser}.}
  \bibinfo{year}{2011}\natexlab{}.
\newblock \showarticletitle{Blended intrinsic maps}.
\newblock \bibinfo{journal}{\emph{ACM transactions on graphics (TOG)}}
  \bibinfo{volume}{30}, \bibinfo{number}{4} (\bibinfo{year}{2011}),
  \bibinfo{pages}{1--12}.
\newblock


\bibitem[\protect\citeauthoryear{Lan, Hong, Yang, Zhou, Meng, Dai, Pan, and
  Loy}{Lan et~al\mbox{.}}{2025a}]%
        {lan2025ln3diff}
\bibfield{author}{\bibinfo{person}{Yushi Lan}, \bibinfo{person}{Fangzhou Hong},
  \bibinfo{person}{Shuai Yang}, \bibinfo{person}{Shangchen Zhou},
  \bibinfo{person}{Xuyi Meng}, \bibinfo{person}{Bo Dai},
  \bibinfo{person}{Xingang Pan}, {and} \bibinfo{person}{Chen~Change Loy}.}
  \bibinfo{year}{2025}\natexlab{a}.
\newblock \showarticletitle{Ln3diff: Scalable latent neural fields diffusion
  for speedy 3d generation}. In \bibinfo{booktitle}{\emph{European Conference
  on Computer Vision}}. Springer, \bibinfo{pages}{112--130}.
\newblock


\bibitem[\protect\citeauthoryear{Lan, Loy, and Dai}{Lan et~al\mbox{.}}{2022}]%
        {lan2022ddf_ijcv}
\bibfield{author}{\bibinfo{person}{Yushi Lan}, \bibinfo{person}{Chen~Change
  Loy}, {and} \bibinfo{person}{Bo Dai}.} \bibinfo{year}{2022}\natexlab{}.
\newblock \showarticletitle{{DDF}: Correspondence Distillation from NeRF-based
  GAN}.
\newblock \bibinfo{journal}{\emph{IJCV}} (\bibinfo{year}{2022}).
\newblock


\bibitem[\protect\citeauthoryear{Lan, Zhou, Lyu, Hong, Yang, Dai, Pan, and
  Loy}{Lan et~al\mbox{.}}{2025b}]%
        {lan2024gaussiananything}
\bibfield{author}{\bibinfo{person}{Yushi Lan}, \bibinfo{person}{Shangchen
  Zhou}, \bibinfo{person}{Zhaoyang Lyu}, \bibinfo{person}{Fangzhou Hong},
  \bibinfo{person}{Shuai Yang}, \bibinfo{person}{Bo Dai},
  \bibinfo{person}{Xingang Pan}, {and} \bibinfo{person}{Chen~Change Loy}.}
  \bibinfo{year}{2025}\natexlab{b}.
\newblock \showarticletitle{GaussianAnything: Interactive Point Cloud Latent
  Diffusion for 3D Generation}. In \bibinfo{booktitle}{\emph{ICLR}}.
\newblock


\bibitem[\protect\citeauthoryear{Liao, Lima, Nehab, Hoppe, Sander, and Yu}{Liao
  et~al\mbox{.}}{2014}]%
        {liao2014automating}
\bibfield{author}{\bibinfo{person}{Jing Liao}, \bibinfo{person}{Rodolfo~S
  Lima}, \bibinfo{person}{Diego Nehab}, \bibinfo{person}{Hugues Hoppe},
  \bibinfo{person}{Pedro~V Sander}, {and} \bibinfo{person}{Jinhui Yu}.}
  \bibinfo{year}{2014}\natexlab{}.
\newblock \showarticletitle{Automating image morphing using structural
  similarity on a halfway domain}.
\newblock \bibinfo{journal}{\emph{ACM Transactions on Graphics (TOG)}}
  \bibinfo{volume}{33}, \bibinfo{number}{5} (\bibinfo{year}{2014}),
  \bibinfo{pages}{1--12}.
\newblock


\bibitem[\protect\citeauthoryear{Lin, Gu, Du, Qu, Chen, Zhang, Gao, Liu, and
  Gunasekaran}{Lin et~al\mbox{.}}{2024}]%
        {lin20242d}
\bibfield{author}{\bibinfo{person}{Jianchu Lin}, \bibinfo{person}{Yinxi Gu},
  \bibinfo{person}{Guangxiao Du}, \bibinfo{person}{Guoqiang Qu},
  \bibinfo{person}{Xiaobing Chen}, \bibinfo{person}{Yudong Zhang},
  \bibinfo{person}{Shangbing Gao}, \bibinfo{person}{Zhen Liu}, {and}
  \bibinfo{person}{Nallappan Gunasekaran}.} \bibinfo{year}{2024}\natexlab{}.
\newblock \showarticletitle{2D/3D Image morphing technology from traditional to
  modern: A survey}.
\newblock \bibinfo{journal}{\emph{Information Fusion}} (\bibinfo{year}{2024}),
  \bibinfo{pages}{102913}.
\newblock


\bibitem[\protect\citeauthoryear{Lipman, Chen, Ben-Hamu, Nickel, and Le}{Lipman
  et~al\mbox{.}}{2022}]%
        {lipman2022flow}
\bibfield{author}{\bibinfo{person}{Yaron Lipman}, \bibinfo{person}{Ricky~TQ
  Chen}, \bibinfo{person}{Heli Ben-Hamu}, \bibinfo{person}{Maximilian Nickel},
  {and} \bibinfo{person}{Matt Le}.} \bibinfo{year}{2022}\natexlab{}.
\newblock \showarticletitle{Flow matching for generative modeling}.
\newblock \bibinfo{journal}{\emph{arXiv preprint arXiv:2210.02747}}
  (\bibinfo{year}{2022}).
\newblock


\bibitem[\protect\citeauthoryear{Mangrulkar, Gugger, Debut, Belkada, Paul, and
  Bossan}{Mangrulkar et~al\mbox{.}}{2022}]%
        {peft}
\bibfield{author}{\bibinfo{person}{Sourab Mangrulkar}, \bibinfo{person}{Sylvain
  Gugger}, \bibinfo{person}{Lysandre Debut}, \bibinfo{person}{Younes Belkada},
  \bibinfo{person}{Sayak Paul}, {and} \bibinfo{person}{Benjamin Bossan}.}
  \bibinfo{year}{2022}\natexlab{}.
\newblock \bibinfo{title}{PEFT: State-of-the-art Parameter-Efficient
  Fine-Tuning methods}.
\newblock \bibinfo{howpublished}{\url{https://github.com/huggingface/peft}}.
\newblock


\bibitem[\protect\citeauthoryear{Michel, Bar-On, Liu, Benaim, and
  Hanocka}{Michel et~al\mbox{.}}{2022}]%
        {michel2022text2mesh}
\bibfield{author}{\bibinfo{person}{Oscar Michel}, \bibinfo{person}{Roi Bar-On},
  \bibinfo{person}{Richard Liu}, \bibinfo{person}{Sagie Benaim}, {and}
  \bibinfo{person}{Rana Hanocka}.} \bibinfo{year}{2022}\natexlab{}.
\newblock \showarticletitle{Text2mesh: Text-driven neural stylization for
  meshes}. In \bibinfo{booktitle}{\emph{Proceedings of the IEEE/CVF Conference
  on Computer Vision and Pattern Recognition}}. \bibinfo{pages}{13492--13502}.
\newblock


\bibitem[\protect\citeauthoryear{Mohammad~Khalid, Xie, Belilovsky, and
  Popa}{Mohammad~Khalid et~al\mbox{.}}{2022}]%
        {mohammad2022clip}
\bibfield{author}{\bibinfo{person}{Nasir Mohammad~Khalid},
  \bibinfo{person}{Tianhao Xie}, \bibinfo{person}{Eugene Belilovsky}, {and}
  \bibinfo{person}{Tiberiu Popa}.} \bibinfo{year}{2022}\natexlab{}.
\newblock \showarticletitle{Clip-mesh: Generating textured meshes from text
  using pretrained image-text models}. In \bibinfo{booktitle}{\emph{SIGGRAPH
  Asia 2022 conference papers}}. \bibinfo{pages}{1--8}.
\newblock


\bibitem[\protect\citeauthoryear{Morreale, Aigerman, Kim, and Mitra}{Morreale
  et~al\mbox{.}}{2024}]%
        {morreale2024neural}
\bibfield{author}{\bibinfo{person}{Luca Morreale}, \bibinfo{person}{Noam
  Aigerman}, \bibinfo{person}{Vladimir~G Kim}, {and} \bibinfo{person}{Niloy~J
  Mitra}.} \bibinfo{year}{2024}\natexlab{}.
\newblock \showarticletitle{Neural semantic surface maps}. In
  \bibinfo{booktitle}{\emph{Computer Graphics Forum}},
  Vol.~\bibinfo{volume}{43}. Wiley Online Library, \bibinfo{pages}{e15005}.
\newblock


\bibitem[\protect\citeauthoryear{OpenAI}{OpenAI}{2023}]%
        {gpt4}
\bibfield{author}{\bibinfo{person}{OpenAI}.} \bibinfo{year}{2023}\natexlab{}.
\newblock \bibinfo{title}{GPT-4: OpenAI's Fourth-Generation Language Model}.
\newblock
\newblock
\urldef\tempurl%
\url{https://openai.com/research/gpt-4}
\showURL{%
\tempurl}


\bibitem[\protect\citeauthoryear{Oquab, Darcet, Moutakanni, Vo, Szafraniec,
  Khalidov, Fernandez, Haziza, Massa, El-Nouby, et~al\mbox{.}}{Oquab
  et~al\mbox{.}}{2023}]%
        {oquab2023dinov2}
\bibfield{author}{\bibinfo{person}{Maxime Oquab}, \bibinfo{person}{Timoth{\'e}e
  Darcet}, \bibinfo{person}{Th{\'e}o Moutakanni}, \bibinfo{person}{Huy Vo},
  \bibinfo{person}{Marc Szafraniec}, \bibinfo{person}{Vasil Khalidov},
  \bibinfo{person}{Pierre Fernandez}, \bibinfo{person}{Daniel Haziza},
  \bibinfo{person}{Francisco Massa}, \bibinfo{person}{Alaaeldin El-Nouby},
  {et~al\mbox{.}}} \bibinfo{year}{2023}\natexlab{}.
\newblock \showarticletitle{Dinov2: Learning robust visual features without
  supervision}.
\newblock \bibinfo{journal}{\emph{arXiv preprint arXiv:2304.07193}}
  (\bibinfo{year}{2023}).
\newblock


\bibitem[\protect\citeauthoryear{Ovsjanikov, Ben-Chen, Solomon, Butscher, and
  Guibas}{Ovsjanikov et~al\mbox{.}}{2012}]%
        {ovsjanikov2012functional}
\bibfield{author}{\bibinfo{person}{Maks Ovsjanikov}, \bibinfo{person}{Mirela
  Ben-Chen}, \bibinfo{person}{Justin Solomon}, \bibinfo{person}{Adrian
  Butscher}, {and} \bibinfo{person}{Leonidas Guibas}.}
  \bibinfo{year}{2012}\natexlab{}.
\newblock \showarticletitle{Functional maps: a flexible representation of maps
  between shapes}.
\newblock \bibinfo{journal}{\emph{ACM Transactions on Graphics (ToG)}}
  \bibinfo{volume}{31}, \bibinfo{number}{4} (\bibinfo{year}{2012}),
  \bibinfo{pages}{1--11}.
\newblock


\bibitem[\protect\citeauthoryear{Peebles and Xie}{Peebles and Xie}{2023}]%
        {peebles2023scalable}
\bibfield{author}{\bibinfo{person}{William Peebles} {and}
  \bibinfo{person}{Saining Xie}.} \bibinfo{year}{2023}\natexlab{}.
\newblock \showarticletitle{Scalable diffusion models with transformers}. In
  \bibinfo{booktitle}{\emph{Proceedings of the IEEE/CVF International
  Conference on Computer Vision}}. \bibinfo{pages}{4195--4205}.
\newblock


\bibitem[\protect\citeauthoryear{Poole, Jain, Barron, and Mildenhall}{Poole
  et~al\mbox{.}}{2022}]%
        {poole2022dreamfusion}
\bibfield{author}{\bibinfo{person}{Ben Poole}, \bibinfo{person}{Ajay Jain},
  \bibinfo{person}{Jonathan~T Barron}, {and} \bibinfo{person}{Ben Mildenhall}.}
  \bibinfo{year}{2022}\natexlab{}.
\newblock \showarticletitle{Dreamfusion: Text-to-3d using 2d diffusion}.
\newblock \bibinfo{journal}{\emph{arXiv preprint arXiv:2209.14988}}
  (\bibinfo{year}{2022}).
\newblock


\bibitem[\protect\citeauthoryear{Radford, Kim, Hallacy, Ramesh, Goh, Agarwal,
  Sastry, Askell, Mishkin, Clark, et~al\mbox{.}}{Radford et~al\mbox{.}}{2021}]%
        {radford2021learning}
\bibfield{author}{\bibinfo{person}{Alec Radford}, \bibinfo{person}{Jong~Wook
  Kim}, \bibinfo{person}{Chris Hallacy}, \bibinfo{person}{Aditya Ramesh},
  \bibinfo{person}{Gabriel Goh}, \bibinfo{person}{Sandhini Agarwal},
  \bibinfo{person}{Girish Sastry}, \bibinfo{person}{Amanda Askell},
  \bibinfo{person}{Pamela Mishkin}, \bibinfo{person}{Jack Clark},
  {et~al\mbox{.}}} \bibinfo{year}{2021}\natexlab{}.
\newblock \showarticletitle{Learning transferable visual models from natural
  language supervision}. In \bibinfo{booktitle}{\emph{International conference
  on machine learning}}. PMLR, \bibinfo{pages}{8748--8763}.
\newblock


\bibitem[\protect\citeauthoryear{Ren, Melzi, Ovsjanikov, and Wonka}{Ren
  et~al\mbox{.}}{2020}]%
        {ren2020maptree}
\bibfield{author}{\bibinfo{person}{Jing Ren}, \bibinfo{person}{Simone Melzi},
  \bibinfo{person}{Maks Ovsjanikov}, {and} \bibinfo{person}{Peter Wonka}.}
  \bibinfo{year}{2020}\natexlab{}.
\newblock \showarticletitle{Maptree: Recovering multiple solutions in the space
  of maps}.
\newblock \bibinfo{journal}{\emph{ACM Transactions on Graphics}}
  \bibinfo{volume}{39}, \bibinfo{number}{6} (\bibinfo{year}{2020}),
  \bibinfo{pages}{1--17}.
\newblock


\bibitem[\protect\citeauthoryear{Rombach, Blattmann, Lorenz, Esser, and
  Ommer}{Rombach et~al\mbox{.}}{2022}]%
        {rombach2022high}
\bibfield{author}{\bibinfo{person}{Robin Rombach}, \bibinfo{person}{Andreas
  Blattmann}, \bibinfo{person}{Dominik Lorenz}, \bibinfo{person}{Patrick
  Esser}, {and} \bibinfo{person}{Bj{\"o}rn Ommer}.}
  \bibinfo{year}{2022}\natexlab{}.
\newblock \showarticletitle{High-resolution image synthesis with latent
  diffusion models}. In \bibinfo{booktitle}{\emph{Proceedings of the IEEE/CVF
  conference on computer vision and pattern recognition}}.
  \bibinfo{pages}{10684--10695}.
\newblock


\bibitem[\protect\citeauthoryear{Saharia, Chan, Saxena, Li, Whang, Denton,
  Ghasemipour, Gontijo~Lopes, Karagol~Ayan, Salimans, et~al\mbox{.}}{Saharia
  et~al\mbox{.}}{2022}]%
        {saharia2022photorealistic}
\bibfield{author}{\bibinfo{person}{Chitwan Saharia}, \bibinfo{person}{William
  Chan}, \bibinfo{person}{Saurabh Saxena}, \bibinfo{person}{Lala Li},
  \bibinfo{person}{Jay Whang}, \bibinfo{person}{Emily~L Denton},
  \bibinfo{person}{Kamyar Ghasemipour}, \bibinfo{person}{Raphael
  Gontijo~Lopes}, \bibinfo{person}{Burcu Karagol~Ayan}, \bibinfo{person}{Tim
  Salimans}, {et~al\mbox{.}}} \bibinfo{year}{2022}\natexlab{}.
\newblock \showarticletitle{Photorealistic text-to-image diffusion models with
  deep language understanding}.
\newblock \bibinfo{journal}{\emph{Advances in neural information processing
  systems}}  \bibinfo{volume}{35} (\bibinfo{year}{2022}),
  \bibinfo{pages}{36479--36494}.
\newblock


\bibitem[\protect\citeauthoryear{Sajjadi, Meyer, Pot, Bergmann, Greff, Radwan,
  Vora, Lu{\v{c}}i{\'c}, Duckworth, Dosovitskiy, et~al\mbox{.}}{Sajjadi
  et~al\mbox{.}}{2022}]%
        {sajjadi2022scene}
\bibfield{author}{\bibinfo{person}{Mehdi~SM Sajjadi}, \bibinfo{person}{Henning
  Meyer}, \bibinfo{person}{Etienne Pot}, \bibinfo{person}{Urs Bergmann},
  \bibinfo{person}{Klaus Greff}, \bibinfo{person}{Noha Radwan},
  \bibinfo{person}{Suhani Vora}, \bibinfo{person}{Mario Lu{\v{c}}i{\'c}},
  \bibinfo{person}{Daniel Duckworth}, \bibinfo{person}{Alexey Dosovitskiy},
  {et~al\mbox{.}}} \bibinfo{year}{2022}\natexlab{}.
\newblock \showarticletitle{Scene representation transformer: Geometry-free
  novel view synthesis through set-latent scene representations}. In
  \bibinfo{booktitle}{\emph{Proceedings of the IEEE/CVF Conference on Computer
  Vision and Pattern Recognition}}. \bibinfo{pages}{6229--6238}.
\newblock


\bibitem[\protect\citeauthoryear{Samuel, Ben-Ari, Darshan, Maron, and
  Chechik}{Samuel et~al\mbox{.}}{2024}]%
        {samuel2024norm}
\bibfield{author}{\bibinfo{person}{Dvir Samuel}, \bibinfo{person}{Rami
  Ben-Ari}, \bibinfo{person}{Nir Darshan}, \bibinfo{person}{Haggai Maron},
  {and} \bibinfo{person}{Gal Chechik}.} \bibinfo{year}{2024}\natexlab{}.
\newblock \showarticletitle{Norm-guided latent space exploration for
  text-to-image generation}.
\newblock \bibinfo{journal}{\emph{Advances in Neural Information Processing
  Systems}}  \bibinfo{volume}{36} (\bibinfo{year}{2024}).
\newblock


\bibitem[\protect\citeauthoryear{Shechtman, Rav-Acha, Irani, and
  Seitz}{Shechtman et~al\mbox{.}}{2010}]%
        {shechtman2010regenerative}
\bibfield{author}{\bibinfo{person}{Eli Shechtman}, \bibinfo{person}{Alex
  Rav-Acha}, \bibinfo{person}{Michal Irani}, {and} \bibinfo{person}{Steve
  Seitz}.} \bibinfo{year}{2010}\natexlab{}.
\newblock \showarticletitle{Regenerative morphing}. In
  \bibinfo{booktitle}{\emph{2010 IEEE Computer Society Conference on Computer
  Vision and Pattern Recognition}}. IEEE, \bibinfo{pages}{615--622}.
\newblock


\bibitem[\protect\citeauthoryear{Shen, Liu, Sun, Ye, Li, Zhang, and Cao}{Shen
  et~al\mbox{.}}{2024}]%
        {shen2024dreammover}
\bibfield{author}{\bibinfo{person}{Liao Shen}, \bibinfo{person}{Tianqi Liu},
  \bibinfo{person}{Huiqiang Sun}, \bibinfo{person}{Xinyi Ye},
  \bibinfo{person}{Baopu Li}, \bibinfo{person}{Jianming Zhang}, {and}
  \bibinfo{person}{Zhiguo Cao}.} \bibinfo{year}{2024}\natexlab{}.
\newblock \showarticletitle{Dreammover: Leveraging the prior of diffusion
  models for image interpolation with large motion}.
\newblock \bibinfo{journal}{\emph{arXiv preprint arXiv:2409.09605}}
  \bibinfo{volume}{2} (\bibinfo{year}{2024}).
\newblock


\bibitem[\protect\citeauthoryear{Shi, Wang, Ye, Long, Li, and Yang}{Shi
  et~al\mbox{.}}{2023}]%
        {shi2023mvdream}
\bibfield{author}{\bibinfo{person}{Yichun Shi}, \bibinfo{person}{Peng Wang},
  \bibinfo{person}{Jianglong Ye}, \bibinfo{person}{Mai Long},
  \bibinfo{person}{Kejie Li}, {and} \bibinfo{person}{Xiao Yang}.}
  \bibinfo{year}{2023}\natexlab{}.
\newblock \showarticletitle{Mvdream: Multi-view diffusion for 3d generation}.
\newblock \bibinfo{journal}{\emph{arXiv preprint arXiv:2308.16512}}
  (\bibinfo{year}{2023}).
\newblock


\bibitem[\protect\citeauthoryear{Si, Huang, Jiang, and Liu}{Si
  et~al\mbox{.}}{2024}]%
        {si2024freeu}
\bibfield{author}{\bibinfo{person}{Chenyang Si}, \bibinfo{person}{Ziqi Huang},
  \bibinfo{person}{Yuming Jiang}, {and} \bibinfo{person}{Ziwei Liu}.}
  \bibinfo{year}{2024}\natexlab{}.
\newblock \showarticletitle{Freeu: Free lunch in diffusion u-net}. In
  \bibinfo{booktitle}{\emph{Proceedings of the IEEE/CVF Conference on Computer
  Vision and Pattern Recognition}}. \bibinfo{pages}{4733--4743}.
\newblock


\bibitem[\protect\citeauthoryear{Solomon, De~Goes, Peyr{\'e}, Cuturi, Butscher,
  Nguyen, Du, and Guibas}{Solomon et~al\mbox{.}}{2015}]%
        {solomon2015convolutional}
\bibfield{author}{\bibinfo{person}{Justin Solomon}, \bibinfo{person}{Fernando
  De~Goes}, \bibinfo{person}{Gabriel Peyr{\'e}}, \bibinfo{person}{Marco
  Cuturi}, \bibinfo{person}{Adrian Butscher}, \bibinfo{person}{Andy Nguyen},
  \bibinfo{person}{Tao Du}, {and} \bibinfo{person}{Leonidas Guibas}.}
  \bibinfo{year}{2015}\natexlab{}.
\newblock \showarticletitle{Convolutional wasserstein distances: Efficient
  optimal transportation on geometric domains}.
\newblock \bibinfo{journal}{\emph{ACM Transactions on Graphics (ToG)}}
  \bibinfo{volume}{34}, \bibinfo{number}{4} (\bibinfo{year}{2015}),
  \bibinfo{pages}{1--11}.
\newblock


\bibitem[\protect\citeauthoryear{Song, Meng, and Ermon}{Song
  et~al\mbox{.}}{2020}]%
        {song2020denoising}
\bibfield{author}{\bibinfo{person}{Jiaming Song}, \bibinfo{person}{Chenlin
  Meng}, {and} \bibinfo{person}{Stefano Ermon}.}
  \bibinfo{year}{2020}\natexlab{}.
\newblock \showarticletitle{Denoising diffusion implicit models}.
\newblock \bibinfo{journal}{\emph{arXiv preprint arXiv:2010.02502}}
  (\bibinfo{year}{2020}).
\newblock


\bibitem[\protect\citeauthoryear{Sorkine and Alexa}{Sorkine and Alexa}{2007}]%
        {sorkine2007rigid}
\bibfield{author}{\bibinfo{person}{Olga Sorkine} {and} \bibinfo{person}{Marc
  Alexa}.} \bibinfo{year}{2007}\natexlab{}.
\newblock \showarticletitle{As-rigid-as-possible surface modeling}. In
  \bibinfo{booktitle}{\emph{Symposium on Geometry processing}},
  Vol.~\bibinfo{volume}{4}. Citeseer, \bibinfo{pages}{109--116}.
\newblock


\bibitem[\protect\citeauthoryear{Sorkine, Cohen-Or, Lipman, Alexa, R{\"o}ssl,
  and Seidel}{Sorkine et~al\mbox{.}}{2004}]%
        {sorkine2004laplacian}
\bibfield{author}{\bibinfo{person}{Olga Sorkine}, \bibinfo{person}{Daniel
  Cohen-Or}, \bibinfo{person}{Yaron Lipman}, \bibinfo{person}{Marc Alexa},
  \bibinfo{person}{Christian R{\"o}ssl}, {and} \bibinfo{person}{H-P Seidel}.}
  \bibinfo{year}{2004}\natexlab{}.
\newblock \showarticletitle{Laplacian surface editing}. In
  \bibinfo{booktitle}{\emph{Proceedings of the 2004 Eurographics/ACM SIGGRAPH
  symposium on Geometry processing}}. \bibinfo{pages}{175--184}.
\newblock


\bibitem[\protect\citeauthoryear{Sun, Guo, Jiang, Mao, Chen, and Huang}{Sun
  et~al\mbox{.}}{2024}]%
        {sun2024srif}
\bibfield{author}{\bibinfo{person}{Mingze Sun}, \bibinfo{person}{Chen Guo},
  \bibinfo{person}{Puhua Jiang}, \bibinfo{person}{Shiwei Mao},
  \bibinfo{person}{Yurun Chen}, {and} \bibinfo{person}{Ruqi Huang}.}
  \bibinfo{year}{2024}\natexlab{}.
\newblock \showarticletitle{SRIF: Semantic Shape Registration Empowered by
  Diffusion-based Image Morphing and Flow Estimation}. In
  \bibinfo{booktitle}{\emph{SIGGRAPH Asia 2024 Conference Papers}}.
  \bibinfo{pages}{1--11}.
\newblock


\bibitem[\protect\citeauthoryear{Tam, Cheng, Lai, Langbein, Liu, Marshall,
  Martin, Sun, and Rosin}{Tam et~al\mbox{.}}{2012}]%
        {tam2012registration}
\bibfield{author}{\bibinfo{person}{Gary~KL Tam}, \bibinfo{person}{Zhi-Quan
  Cheng}, \bibinfo{person}{Yu-Kun Lai}, \bibinfo{person}{Frank~C Langbein},
  \bibinfo{person}{Yonghuai Liu}, \bibinfo{person}{David Marshall},
  \bibinfo{person}{Ralph~R Martin}, \bibinfo{person}{Xian-Fang Sun}, {and}
  \bibinfo{person}{Paul~L Rosin}.} \bibinfo{year}{2012}\natexlab{}.
\newblock \showarticletitle{Registration of 3D point clouds and meshes: A
  survey from rigid to nonrigid}.
\newblock \bibinfo{journal}{\emph{IEEE transactions on visualization and
  computer graphics}} \bibinfo{volume}{19}, \bibinfo{number}{7}
  (\bibinfo{year}{2012}), \bibinfo{pages}{1199--1217}.
\newblock


\bibitem[\protect\citeauthoryear{Tang, Jia, Wang, Phoo, and Hariharan}{Tang
  et~al\mbox{.}}{2023}]%
        {tang2023emergent}
\bibfield{author}{\bibinfo{person}{Luming Tang}, \bibinfo{person}{Menglin Jia},
  \bibinfo{person}{Qianqian Wang}, \bibinfo{person}{Cheng~Perng Phoo}, {and}
  \bibinfo{person}{Bharath Hariharan}.} \bibinfo{year}{2023}\natexlab{}.
\newblock \showarticletitle{Emergent correspondence from image diffusion}.
\newblock \bibinfo{journal}{\emph{Advances in Neural Information Processing
  Systems}}  \bibinfo{volume}{36} (\bibinfo{year}{2023}),
  \bibinfo{pages}{1363--1389}.
\newblock


\bibitem[\protect\citeauthoryear{Tsai, Sun, and Chen}{Tsai
  et~al\mbox{.}}{2022}]%
        {tsai2022multiview}
\bibfield{author}{\bibinfo{person}{Chih-Jung Tsai}, \bibinfo{person}{Cheng
  Sun}, {and} \bibinfo{person}{Hwann-Tzong Chen}.}
  \bibinfo{year}{2022}\natexlab{}.
\newblock \showarticletitle{Multiview Regenerative Morphing with Dual Flows}.
  In \bibinfo{booktitle}{\emph{European Conference on Computer Vision}}.
  Springer, \bibinfo{pages}{492--509}.
\newblock


\bibitem[\protect\citeauthoryear{Vahdat, Williams, Gojcic, Litany, Fidler,
  Kreis, et~al\mbox{.}}{Vahdat et~al\mbox{.}}{2022}]%
        {vahdat2022lion}
\bibfield{author}{\bibinfo{person}{Arash Vahdat}, \bibinfo{person}{Francis
  Williams}, \bibinfo{person}{Zan Gojcic}, \bibinfo{person}{Or Litany},
  \bibinfo{person}{Sanja Fidler}, \bibinfo{person}{Karsten Kreis},
  {et~al\mbox{.}}} \bibinfo{year}{2022}\natexlab{}.
\newblock \showarticletitle{Lion: Latent point diffusion models for 3d shape
  generation}.
\newblock \bibinfo{journal}{\emph{Advances in Neural Information Processing
  Systems}}  \bibinfo{volume}{35} (\bibinfo{year}{2022}),
  \bibinfo{pages}{10021--10039}.
\newblock


\bibitem[\protect\citeauthoryear{Vaswani}{Vaswani}{2017}]%
        {vaswani2017attention}
\bibfield{author}{\bibinfo{person}{A Vaswani}.}
  \bibinfo{year}{2017}\natexlab{}.
\newblock \showarticletitle{Attention is all you need}.
\newblock \bibinfo{journal}{\emph{Advances in Neural Information Processing
  Systems}} (\bibinfo{year}{2017}).
\newblock


\bibitem[\protect\citeauthoryear{Vyas, Chen, Mohanty, Jiang, and
  Krishnamurthy}{Vyas et~al\mbox{.}}{2021}]%
        {vyas2021latent}
\bibfield{author}{\bibinfo{person}{Shantanu Vyas}, \bibinfo{person}{Ting-Ju
  Chen}, \bibinfo{person}{Ronak~R Mohanty}, \bibinfo{person}{Peng Jiang}, {and}
  \bibinfo{person}{Vinayak~R Krishnamurthy}.} \bibinfo{year}{2021}\natexlab{}.
\newblock \showarticletitle{Latent embedded graphs for image and shape
  interpolation}.
\newblock \bibinfo{journal}{\emph{Computer-Aided Design}}
  \bibinfo{volume}{140} (\bibinfo{year}{2021}), \bibinfo{pages}{103091}.
\newblock


\bibitem[\protect\citeauthoryear{Wolberg}{Wolberg}{1998}]%
        {wolberg1998image}
\bibfield{author}{\bibinfo{person}{George Wolberg}.}
  \bibinfo{year}{1998}\natexlab{}.
\newblock \showarticletitle{Image morphing: a survey}.
\newblock \bibinfo{journal}{\emph{The visual computer}} \bibinfo{volume}{14},
  \bibinfo{number}{8-9} (\bibinfo{year}{1998}), \bibinfo{pages}{360--372}.
\newblock


\bibitem[\protect\citeauthoryear{Wu, Si, Jiang, Huang, and Liu}{Wu
  et~al\mbox{.}}{2025}]%
        {wu2025freeinit}
\bibfield{author}{\bibinfo{person}{Tianxing Wu}, \bibinfo{person}{Chenyang Si},
  \bibinfo{person}{Yuming Jiang}, \bibinfo{person}{Ziqi Huang}, {and}
  \bibinfo{person}{Ziwei Liu}.} \bibinfo{year}{2025}\natexlab{}.
\newblock \showarticletitle{Freeinit: Bridging initialization gap in video
  diffusion models}. In \bibinfo{booktitle}{\emph{European Conference on
  Computer Vision}}. Springer, \bibinfo{pages}{378--394}.
\newblock


\bibitem[\protect\citeauthoryear{Xiang, Lv, Xu, Deng, Wang, Zhang, Chen, Tong,
  and Yang}{Xiang et~al\mbox{.}}{2024}]%
        {xiang2024structured}
\bibfield{author}{\bibinfo{person}{Jianfeng Xiang}, \bibinfo{person}{Zelong
  Lv}, \bibinfo{person}{Sicheng Xu}, \bibinfo{person}{Yu Deng},
  \bibinfo{person}{Ruicheng Wang}, \bibinfo{person}{Bowen Zhang},
  \bibinfo{person}{Dong Chen}, \bibinfo{person}{Xin Tong}, {and}
  \bibinfo{person}{Jiaolong Yang}.} \bibinfo{year}{2024}\natexlab{}.
\newblock \showarticletitle{Structured 3D Latents for Scalable and Versatile 3D
  Generation}.
\newblock \bibinfo{journal}{\emph{arXiv preprint arXiv:2412.01506}}
  (\bibinfo{year}{2024}).
\newblock


\bibitem[\protect\citeauthoryear{Xu, Cheng, Gao, Wang, Gao, and Shan}{Xu
  et~al\mbox{.}}{2024}]%
        {xu2024instantmesh}
\bibfield{author}{\bibinfo{person}{Jiale Xu}, \bibinfo{person}{Weihao Cheng},
  \bibinfo{person}{Yiming Gao}, \bibinfo{person}{Xintao Wang},
  \bibinfo{person}{Shenghua Gao}, {and} \bibinfo{person}{Ying Shan}.}
  \bibinfo{year}{2024}\natexlab{}.
\newblock \showarticletitle{Instantmesh: Efficient 3d mesh generation from a
  single image with sparse-view large reconstruction models}.
\newblock \bibinfo{journal}{\emph{arXiv preprint arXiv:2404.07191}}
  (\bibinfo{year}{2024}).
\newblock


\bibitem[\protect\citeauthoryear{Yang, Wang, Lan, Fan, Peng, Yang, and
  Dong}{Yang et~al\mbox{.}}{2024}]%
        {yang2024learning}
\bibfield{author}{\bibinfo{person}{Songlin Yang}, \bibinfo{person}{Wei Wang},
  \bibinfo{person}{Yushi Lan}, \bibinfo{person}{Xiangyu Fan},
  \bibinfo{person}{Bo Peng}, \bibinfo{person}{Lei Yang}, {and}
  \bibinfo{person}{Jing Dong}.} \bibinfo{year}{2024}\natexlab{}.
\newblock \showarticletitle{Learning dense correspondence for nerf-based face
  reenactment}. In \bibinfo{booktitle}{\emph{Proceedings of the AAAI Conference
  on Artificial Intelligence}}, Vol.~\bibinfo{volume}{38}.
  \bibinfo{pages}{6522--6530}.
\newblock


\bibitem[\protect\citeauthoryear{Yang, Wang, Ling, Peng, Tan, and Dong}{Yang
  et~al\mbox{.}}{2023a}]%
        {yang2023context}
\bibfield{author}{\bibinfo{person}{Songlin Yang}, \bibinfo{person}{Wei Wang},
  \bibinfo{person}{Jun Ling}, \bibinfo{person}{Bo Peng}, \bibinfo{person}{Xu
  Tan}, {and} \bibinfo{person}{Jing Dong}.} \bibinfo{year}{2023}\natexlab{a}.
\newblock \showarticletitle{Context-aware talking-head video editing}. In
  \bibinfo{booktitle}{\emph{Proceedings of the 31st ACM International
  Conference on Multimedia}}. \bibinfo{pages}{7718--7727}.
\newblock


\bibitem[\protect\citeauthoryear{Yang, Wang, Peng, and Dong}{Yang
  et~al\mbox{.}}{2023b}]%
        {yang2023designing}
\bibfield{author}{\bibinfo{person}{Songlin Yang}, \bibinfo{person}{Wei Wang},
  \bibinfo{person}{Bo Peng}, {and} \bibinfo{person}{Jing Dong}.}
  \bibinfo{year}{2023}\natexlab{b}.
\newblock \showarticletitle{Designing a 3D-aware StyleNeRF encoder for face
  editing}. In \bibinfo{booktitle}{\emph{ICASSP 2023-2023 IEEE International
  Conference on Acoustics, Speech and Signal Processing (ICASSP)}}. IEEE,
  \bibinfo{pages}{1--5}.
\newblock


\bibitem[\protect\citeauthoryear{Yu, Kwak, Jang, Jeong, Huang, Shin, and
  Xie}{Yu et~al\mbox{.}}{2024}]%
        {yu2024representation}
\bibfield{author}{\bibinfo{person}{Sihyun Yu}, \bibinfo{person}{Sangkyung
  Kwak}, \bibinfo{person}{Huiwon Jang}, \bibinfo{person}{Jongheon Jeong},
  \bibinfo{person}{Jonathan Huang}, \bibinfo{person}{Jinwoo Shin}, {and}
  \bibinfo{person}{Saining Xie}.} \bibinfo{year}{2024}\natexlab{}.
\newblock \showarticletitle{Representation alignment for generation: Training
  diffusion transformers is easier than you think}.
\newblock \bibinfo{journal}{\emph{arXiv preprint arXiv:2410.06940}}
  (\bibinfo{year}{2024}).
\newblock


\bibitem[\protect\citeauthoryear{Yumer, Chaudhuri, Hodgins, and Kara}{Yumer
  et~al\mbox{.}}{2015}]%
        {yumer2015semantic}
\bibfield{author}{\bibinfo{person}{Mehmet~Ersin Yumer},
  \bibinfo{person}{Siddhartha Chaudhuri}, \bibinfo{person}{Jessica~K Hodgins},
  {and} \bibinfo{person}{Levent~Burak Kara}.} \bibinfo{year}{2015}\natexlab{}.
\newblock \showarticletitle{Semantic shape editing using deformation handles}.
\newblock \bibinfo{journal}{\emph{ACM Transactions on Graphics (TOG)}}
  \bibinfo{volume}{34}, \bibinfo{number}{4} (\bibinfo{year}{2015}),
  \bibinfo{pages}{1--12}.
\newblock


\bibitem[\protect\citeauthoryear{Zhan, Fu, and Ritchie}{Zhan
  et~al\mbox{.}}{2024}]%
        {zhan2024charactermixer}
\bibfield{author}{\bibinfo{person}{Xiao Zhan}, \bibinfo{person}{Rao Fu}, {and}
  \bibinfo{person}{Daniel Ritchie}.} \bibinfo{year}{2024}\natexlab{}.
\newblock \showarticletitle{CharacterMixer: Rig-Aware Interpolation of 3D
  Characters}. In \bibinfo{booktitle}{\emph{Computer Graphics Forum}}. Wiley
  Online Library, \bibinfo{pages}{e15047}.
\newblock


\bibitem[\protect\citeauthoryear{Zhang, Tang, Niessner, and Wonka}{Zhang
  et~al\mbox{.}}{2023}]%
        {zhang20233dshape2vecset}
\bibfield{author}{\bibinfo{person}{Biao Zhang}, \bibinfo{person}{Jiapeng Tang},
  \bibinfo{person}{Matthias Niessner}, {and} \bibinfo{person}{Peter Wonka}.}
  \bibinfo{year}{2023}\natexlab{}.
\newblock \showarticletitle{3dshape2vecset: A 3d shape representation for
  neural fields and generative diffusion models}.
\newblock \bibinfo{journal}{\emph{ACM Transactions on Graphics (TOG)}}
  \bibinfo{volume}{42}, \bibinfo{number}{4} (\bibinfo{year}{2023}),
  \bibinfo{pages}{1--16}.
\newblock


\bibitem[\protect\citeauthoryear{Zhang, Zhou, Xu, Dai, and Pan}{Zhang
  et~al\mbox{.}}{2024}]%
        {zhang2024diffmorpher}
\bibfield{author}{\bibinfo{person}{Kaiwen Zhang}, \bibinfo{person}{Yifan Zhou},
  \bibinfo{person}{Xudong Xu}, \bibinfo{person}{Bo Dai}, {and}
  \bibinfo{person}{Xingang Pan}.} \bibinfo{year}{2024}\natexlab{}.
\newblock \showarticletitle{DiffMorpher: Unleashing the Capability of Diffusion
  Models for Image Morphing}. In \bibinfo{booktitle}{\emph{Proceedings of the
  IEEE/CVF Conference on Computer Vision and Pattern Recognition}}.
  \bibinfo{pages}{7912--7921}.
\newblock


\bibitem[\protect\citeauthoryear{Zhang, Xu, Peng, Rahmani, and Liu}{Zhang
  et~al\mbox{.}}{2025}]%
        {zhang2025diff}
\bibfield{author}{\bibinfo{person}{Zhengbo Zhang}, \bibinfo{person}{Li Xu},
  \bibinfo{person}{Duo Peng}, \bibinfo{person}{Hossein Rahmani}, {and}
  \bibinfo{person}{Jun Liu}.} \bibinfo{year}{2025}\natexlab{}.
\newblock \showarticletitle{Diff-tracker: text-to-image diffusion models are
  unsupervised trackers}. In \bibinfo{booktitle}{\emph{European Conference on
  Computer Vision}}. Springer, \bibinfo{pages}{319--337}.
\newblock


\bibitem[\protect\citeauthoryear{Zhu, Ju, Zhang, Wang, Yuan, Hu, and Xu}{Zhu
  et~al\mbox{.}}{2024}]%
        {zhu2024densematcher}
\bibfield{author}{\bibinfo{person}{Junzhe Zhu}, \bibinfo{person}{Yuanchen Ju},
  \bibinfo{person}{Junyi Zhang}, \bibinfo{person}{Muhan Wang},
  \bibinfo{person}{Zhecheng Yuan}, \bibinfo{person}{Kaizhe Hu}, {and}
  \bibinfo{person}{Huazhe Xu}.} \bibinfo{year}{2024}\natexlab{}.
\newblock \showarticletitle{DenseMatcher: Learning 3D Semantic Correspondence
  for Category-Level Manipulation from a Single Demo}.
\newblock \bibinfo{journal}{\emph{arXiv preprint arXiv:2412.05268}}
  (\bibinfo{year}{2024}).
\newblock


\bibitem[\protect\citeauthoryear{Zope and Zope}{Zope and Zope}{2017}]%
        {zope2017survey}
\bibfield{author}{\bibinfo{person}{Bhushan Zope} {and}
  \bibinfo{person}{Soniya~B Zope}.} \bibinfo{year}{2017}\natexlab{}.
\newblock \showarticletitle{A Survey of Morphing Techniques}.
\newblock \bibinfo{journal}{\emph{International Journal of Advanced
  Engineering, Management and Science}} \bibinfo{volume}{3},
  \bibinfo{number}{2} (\bibinfo{year}{2017}), \bibinfo{pages}{239773}.
\newblock


\end{thebibliography}

\newpage
\appendix

\section{Supplementary Materials}

\subsection{Outline}

To experimentally validate the rationale behind the motivations discussed in our manuscript and to provide additional details that could not be elaborated on due to manuscript space limitations, we have carefully prepared comprehensive supplementary materials for reference (\textbf{\textit{Click on the index to directly access the corresponding content}}).

\begin{itemize}
    \item [\ref{SM: ga}] 3D Generation Model: Gaussian Anything
    \item [\ref{SM: align}] How to Align/Prepare the Input 3D and Images with Gaussian Anything?
    \item [\ref{SM: 3D}] How to Choose Appropriate 3D Diffusion Models for 3D Morphing?
    \item [\ref{SM: scale}] Effects of Scale in the Low-Frequency Enhancement
    \item [\ref{SM: baseline}] Baseline Methods and Implementation Details
    \item [\ref{SM: shape}] Comparisons with Shape Morphing Methods
    \item [\ref{SM: explicit}] Exploratory Experiments with Explicit Correspondence
    \item [\ref{SM: testing}] Testing Protocol and Cases
    \item [\ref{SM: user study}] Raw Statistics of User Study
\end{itemize}

\subsection{3D Generation Model: Gaussian Anything}
\label{SM: ga}

Gaussian Anything~\cite{lan2024gaussiananything} introduces a 3D generation framework built on a point cloud-based 3D latent space. The 3D Variational Autoencoder (VAE) (See~\ref{VAE}) efficiently encodes 3D data into a dynamic latent space, which is subsequently decoded into detailed Surfel Gaussians. Diffusion models (See~\ref{Diffusion}) trained on this compacted latent space achieve remarkable results in 3D generation and editing conditioned on text, as well as in generating high-quality 3D content from images on diverse real-world datasets. For more implementation details, please see their \href{https://nirvanalan.github.io/projects/GA/}{\textcolor{red}{project page}}.

\subsubsection{Point-Cloud Structured 3D VAE} 
\label{VAE}

A 3D VAE is introduced that takes multi-view posed RGB-D (Depth)-Normal renderings as input. These renderings are easy to generate and provide a rich set of 3D attributes corresponding to the input object. Each view's information is concatenated along the channel dimension and efficiently encoded using a scene representation transformer~\cite{sajjadi2022scene}, producing a compact latent representation of the 3D input. Rather than directly applying this latent representation to diffusion learning, the model's innovative method transforms unordered tokens into a shape that mirrors the 3D input. This transformation is achieved by cross-attending~\cite{huang2024pointinfinity} the latent set with a sparse point cloud sampled from the 3D shape. This point-cloud structured latent space significantly aids in disentangling shape and texture, as well as enabling 3D editing. Subsequently, a DiT-based 3D decoder~\cite{peebles2023scalable} progressively decodes and upscales the latent point cloud into a dense set of Surfel Gaussians~\cite{huang20242d}, which are rasterized into high-resolution renderings to guide the 3D VAE training.

\begin{figure}[t]
    \centering
    \includegraphics[width=0.49\textwidth]{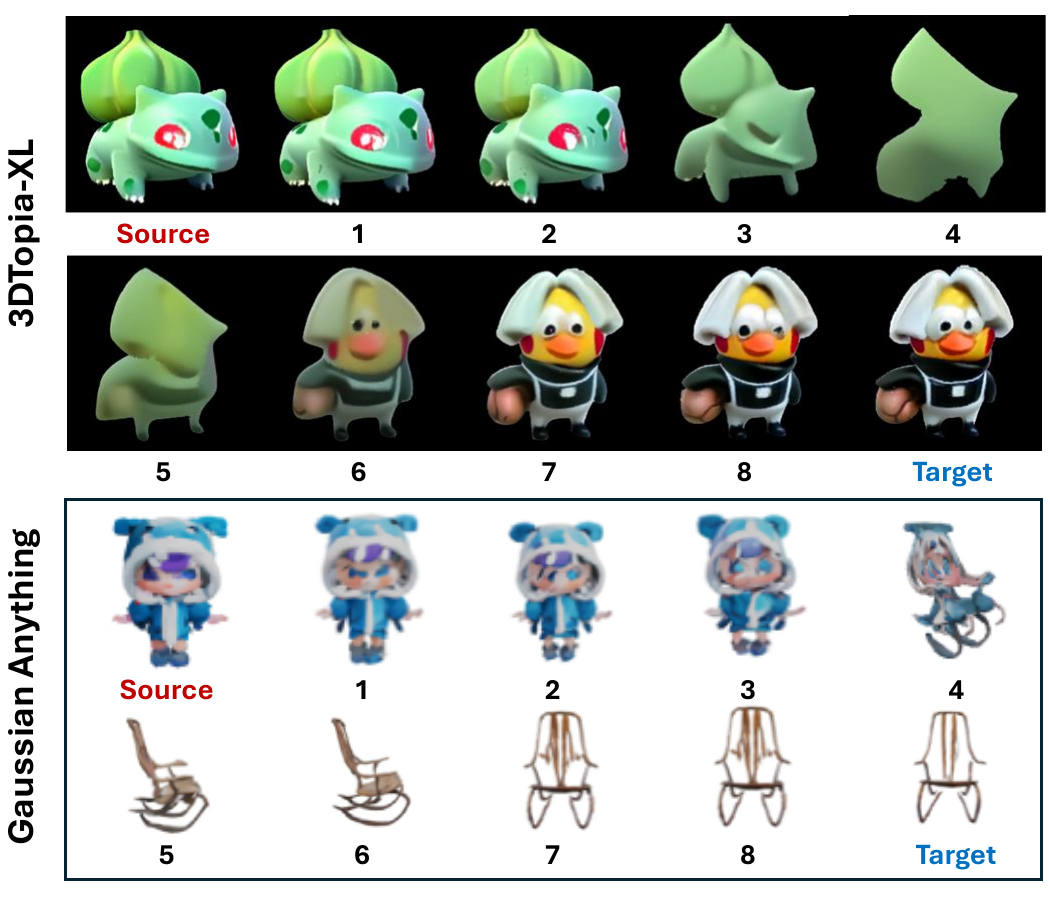}
    \caption{Evaluation of 3D generation model capabilities. Based on accessibility, we tested the interpolation performance of projects with available training code and model details, namely 3DTopia-XL~\cite{chen20243dtopia} and Gaussian Anything~\cite{lan2024gaussiananything} in the image-to-3D setting. We found that while 3DTopia-XL generates high-quality 3D assets, its latent space lacks reasonable generative capabilities, as evidenced by the interpolation results between the 3rd and 6th samples.}
    \label{fig: 3d_generation_model}
\end{figure}

\subsubsection{Cascaded 3D Generation with Flow Matching}
\label{Diffusion}

After the 3D VAE is trained, they conduct cascaded latent diffusion modeling on the latent space through flow matching~\cite{albergo2023stochastic} using the DiT~\cite{peebles2023scalable} framework. To encourage better shape-texture disentanglement, a point cloud diffusion model is first trained to carve the overall layout of the input shape. Then, a point cloud feature diffusion model is cascaded to output the corresponding feature conditioned on the generated point cloud. The generated featured point cloud is then decoded into Surfel Gaussians~\cite{huang20242d} via pre-trained VAE for downstream applications.

\subsection{How to Align/Prepare the Input 3D and Images with Gaussian Anything?}
\label{SM: align}

For textured 3D representations, multi-view RGB, depth, and normal images can be directly rendered, and then the corresponding latent can be obtained using the 3D VAE of Gaussian Anything. For a single image, two methods are possible: (a) The 2D image can be lifted to multi-view using a multi-view generation model~\cite{shi2023mvdream}, and then a renderable textured 3D model can be trained from these multi-view images, or (b) A direct image-to-3D method~\cite{huang2024mvadapter} can be used to obtain the textured 3D model.

\begin{figure*}[t]
    \centering
    \includegraphics[width=\textwidth]{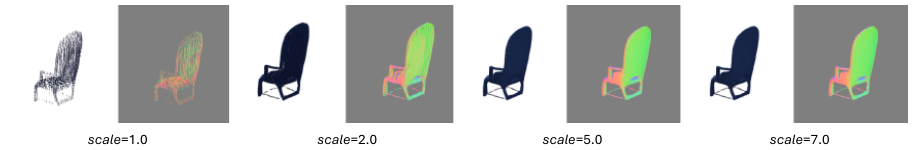}
    \caption{The effects of $scale$ in the Low-Frequency Enhancement.}
    \label{fig: scale}
\end{figure*}

\subsection{How to Choose Appropriate 3D Diffusion Models for 3D Morphing?}
\label{SM: 3D}

Selecting an appropriate 3D generative model is foundational for texured 3D regenerative morphing, as it determines (a) the range of 3D object categories that can be handled and (b) the ability to integrate diverse information for generating smooth interpolation sequences. We followed four criteria when selecting a 3D generative model for our research:  

\textbf{(a) Accessibility:} Training 3D generative models is highly resource-intensive, and high-quality models capable of generating diverse outputs are often proprietary assets, accessible only through APIs. This limits our ability to probe the internal characteristics and potential issues of such models. Therefore, an ideal 3D generative model should be open-sourced, including all testing and training files, datasets, and model checkpoints. Based on this criterion, we selected Gaussian Anything~\cite{lan2024gaussiananything} and 3DTopia-XL~\cite{chen20243dtopia} as our potential target models.  

\textbf{(b) Fairness}: For data-driven studies, fairness is reflected in the use of publicly available datasets for training, ensuring future researchers can build on our work with a well-established baseline or benchmark. The Objaverse~\cite{deitke2023objaverse} dataset, currently one of the most widely adopted 3D datasets, is particularly suitable for academic research. Thus, we prefer Gaussian Anything~\cite{lan2024gaussiananything}, which is trained on Objaverse.  

\textbf{(c) Generation Quality}: 3D generative modeling has become one of the most competitive and rapidly evolving research areas in recent years, with many papers showcasing impressive results. However, unlike 2D images or videos, 3D training data is harder to collect at scale, and no existing model can perfectly generate a full range of 3D objects. Therefore, we prioritized models capable of generating a wide variety of objects, ideally including categories such as animals, buildings, furniture, food, transportation, and plants. After evaluating the performance of several state-of-the-art 3D generative models on image-to-3D and text-to-3D tasks, we selected Gaussian Anything~\cite{lan2024gaussiananything} as our research model.  

\textbf{(d) Preliminary Interpolation Feasibility}: For models with strong generative capabilities, the quickest way to determine their suitability for 3D morphing research is to conduct basic interpolation tests. Not all generative models can effectively fuse different information for interpolation. Among the tested models (as shown in Fig.~\ref{fig: 3d_generation_model}), Gaussian Anything~\cite{lan2024gaussiananything} demonstrated the best interpolation performance, making it the most suitable choice for our study. \vspace{-0.2cm}

\subsection{Effects of Scale in the Low-Frequency Enhancement}
\label{SM: scale}

As shown in Fig.~\ref{fig: scale}, increasing the scale continuously enhances the surface generation capability. However, through empirical observation, we found that beyond a certain value, the surface does not increase further with larger scale values. Therefore, setting the scale to 5 is optimal.

\subsection{Baseline Methods and Implementation Details}
\label{SM: baseline}

\subsubsection{DiffMorpher}

Given two images, DiffMorpher~\cite{zhang2024diffmorpher} uses two LoRAs~\cite{hu2021lora} to fit the two images respectively. Then the latent noises for the two images are obtained via DDIM inversion~\cite{song2020denoising}. The mean and standard deviation of the interpolated noises are adjusted through AdaIN. To generate an intermediate image, they interpolate between both the LoRA parameters and the latent noises via the interpolation ratio $\alpha$. In addition, the text embedding and the K and V in self-attention modules are also replaced with the interpolation between the corresponding components. Using a sequence of $\alpha$ and a new sampling schedule, their method will produce a series of high-fidelity images depicting a smooth transition between the two given images. We followed the script and default parameter settings given by Diffmorpher and used their \href{https://github.com/Kevin-thu/DiffMorpher}{\textcolor{red}{open-source code}} to produce the results.

\subsubsection{AID}

Similar to the DiffMorpher~\cite{zhang2024diffmorpher} framework, AID~\cite{he2024aid} removes the LoRA fitting and introduces the following additional modifications: (a) Replacing both cross-attention and self-attention mechanisms during interpolated image generation with fused interpolated attention; (b) Selecting interpolation coefficients using a Beta prior; (c) Injecting prompt guidance into the fused interpolated cross-attention. We implemented the generation of relevant results based on the code of Stable Diffusion 1.5~\cite{rombach2022high}, and all settings follow the default settings of AID. More details can be found on their \href{https://github.com/QY-H00/attention-interpolation-diffusion}{\textcolor{red}{project page}}.

\subsubsection{MV-Adapter}

MV-Adapter~\cite{huang2024mvadapter} is a versatile plug-and-play and state-of-the-art adapter that turns existing pre-trained text-to-image (T2I) diffusion models to multi-view image generators. We generated image morphing results based on their Image-to-Multiview \href{https://github.com/huanngzh/MV-Adapter}{\textcolor{red}{code}} and Stable Diffusion 2.1. The only change is that we linearly interpolated the condition features of the source image and target image extracted by their image encoder according to different morphing weights.

\begin{figure*}[t]
    \centering
    \includegraphics[width=1.0\textwidth]{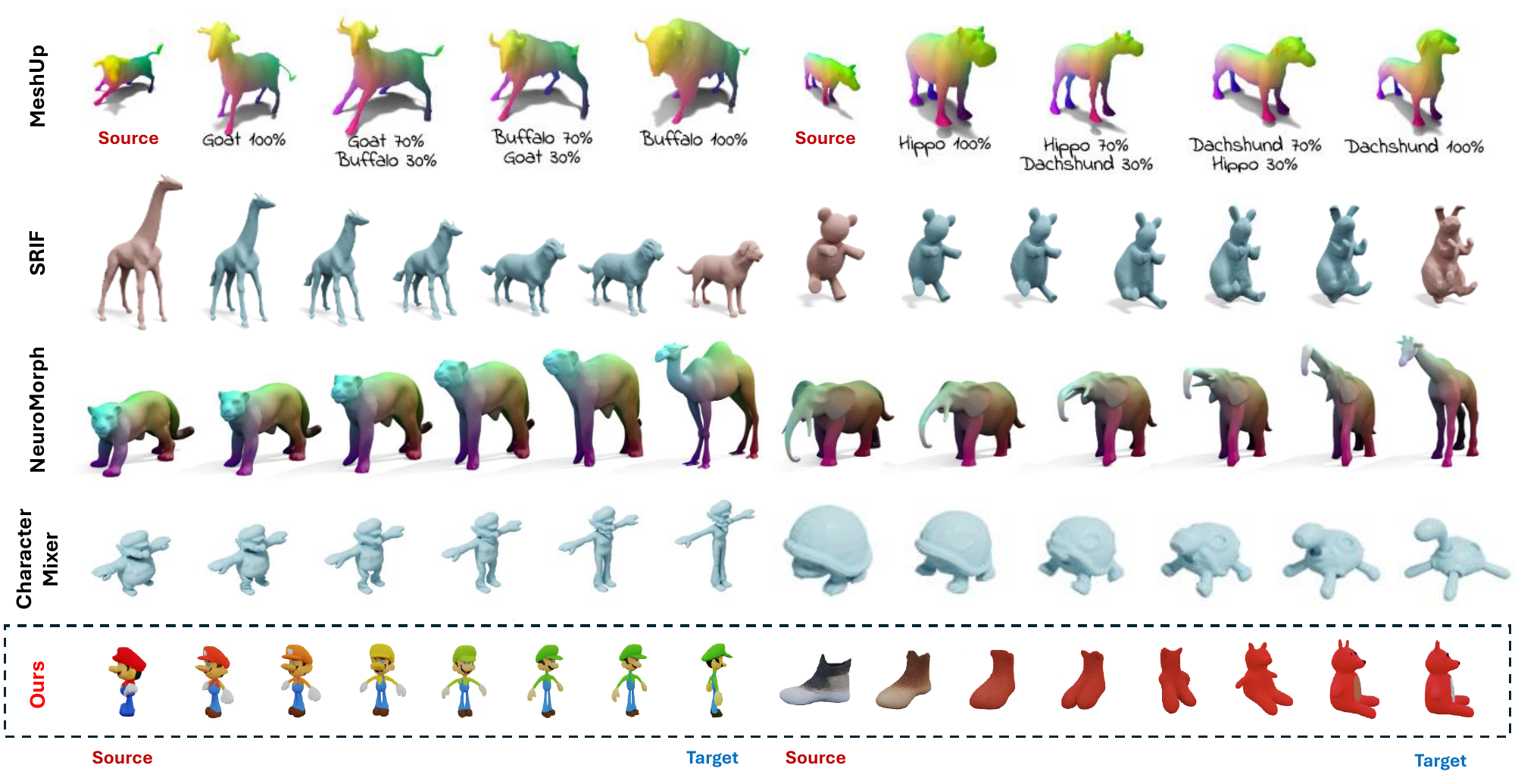}
 
    \caption{Comparison of related methods. Our method focuses on textured 3D morphing, whereas MeshUp~\cite{kim2024meshup}, SRIF~\cite{sun2024srif}, NeuroMorph~\cite{eisenberger2021neuromorph}, and CharacterMixer~\cite{zhan2024charactermixer} are limited to shape-only 3D morphing. Note that all results outside the dashed boxes are sourced from their respective manuscripts.}
    \label{fig: shape morphing}
\end{figure*}

\subsubsection{Luma}

The \href{https://dream-machine.lumalabs.ai/}{\textcolor{red}{Dream Machine}} of Luma AI is based on the DiT~\cite{peebles2023scalable} video generation architecture, capable of generating high-quality videos with 120 frames in just 120 seconds, enabling rapid creative iteration. It understands physical interactions, ensuring that the generated video characters and scenes maintain consistency and physical accuracy. We accessed their API and utilized the video generation function to generate intermediate video frames by providing the source image as the first frame and the target image as the last frame. For instance, for the "polar bear" to "wooden stool" morphing video generation, the guiding prompt we used is: \textit{"Morph a polar bear into a wooden stool, smoothly interpolating both geometry and texture, with the object always remaining at the center of the frame."}

\subsubsection{MorphFlow}

MorphFlow~\cite{tsai2022multiview} introduces an optimization-based method for multi-view regenerative morphing. The method does not assume prior knowledge of the categories or affinities between the source and target images, nor does it rely on predefined correspondences. By utilizing optimal transport, the method interpolates a volume for rendering smooth multi-view transitions. Additionally, a rigid transformation is incorporated to preserve structure during the morphing process. The method is highly efficient, learning and rendering a morphing renderer from scratch in just 30 minutes, with the ability to generate a novel-view morph per second during morphing and rendering. We first obtain the multi-view images along with their corresponding COLMAP camera annotations, and then generate the morphing output using their \href{https://github.com/jimtsai23/MorphFlow}{\textcolor{red}{open-source code}} with the default parameter settings.

\begin{table}[t]
\footnotesize
    \centering
    \caption{Task setting comparison.}
    \vspace{-0.4cm}
    \begin{tabular}{ccccc}
    \toprule
    & Shape  & Texture & Aligned Dataset  & Out-of-Domain Morphing \\ \midrule
MeshUp & {\color{green} \checkmark} & {\color{red} $\times$} & {\color{green} No Need} & {\color{green} \checkmark} \\ 
SRIF & {\color{green} \checkmark} & {\color{red} $\times$} & {\color{green} No Need} & {\color{green} \checkmark} \\ 
NeuroMorph & {\color{green} \checkmark} & {\color{red} $\times$} & {\color{red} Need} & {\color{red} $\times$} \\ 
CharacterMixer& {\color{green} \checkmark} & {\color{red} $\times$} & {\color{red} Need}  & {\color{red} $\times$} \\ 
\textbf{MorphFlow} & {\color{green} \checkmark} & {\color{green} \checkmark} & {\color{green} No Need}  & {\color{green} \checkmark}\\
\textbf{Ours} & {\color{green} \checkmark} & {\color{green} \checkmark} & {\color{green} No Need}  & {\color{green} \checkmark} \\
\bottomrule
    \end{tabular}
    
    \label{tab: shape morphing}
    \vspace{-0.5cm}
\end{table}

\subsection{Comparisons with Shape Morphing Methods}
\label{SM: shape}

As shown in Tab.~\ref{tab: shape morphing}, our setting focuses on textured 3D morphing, a task currently shared only with MorphFlow~\cite{tsai2022multiview}. Earlier works like NeuroMorph~\cite{eisenberger2021neuromorph} and CharacterMixer~\cite{zhan2024charactermixer} were trained on aligned datasets, essentially learning in-domain, topology-aligned correspondences between 3D data. However, these methods fail to generalize such correspondences to out-of-domain 3D representations. Other methods, such as MeshUp~\cite{kim2024meshup} and SRIF~\cite{sun2024srif}, explored shape morphing by leveraging generative priors. While they recognized the importance of generative priors for improving generalization in morphing tasks, their work was limited to shape-only morphing and did not release source code. We qualitatively compared results from their manuscripts with ours, demonstrating that our method not only performs morphing between textured 3D representations with similar topologies (e.g., Mario and Luigi) but also handles morphing between representations with significant category differences (e.g., a boot and a teddy bear).

\begin{figure*}[t]
    \centering
    \includegraphics[width=0.73\textwidth]{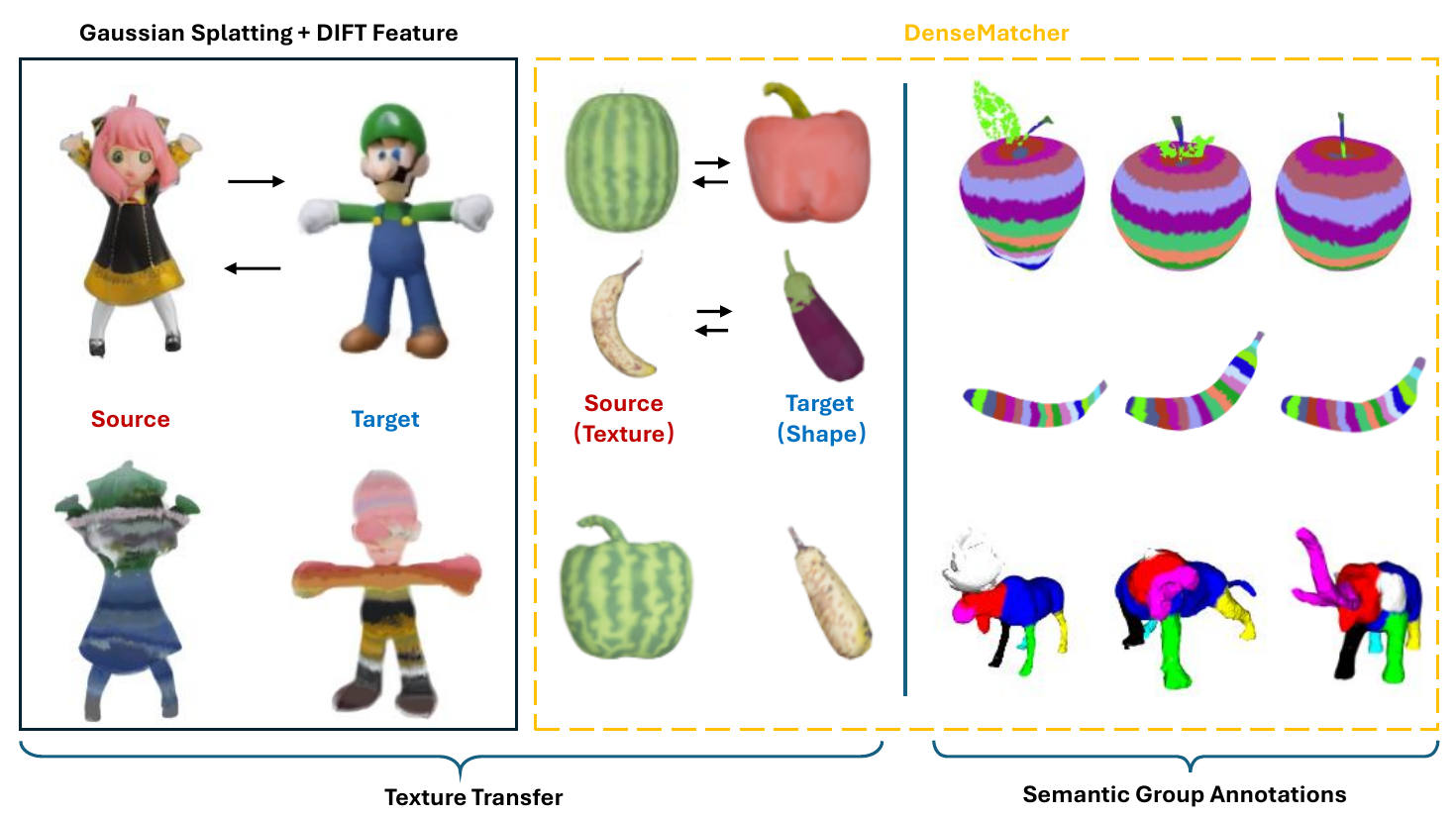}
    \caption{Exploring possibilities based on explicit correspondence. Using explicit correspondence for morphing presents two major challenges: first, obtaining semantic features for tens of thousands of points is extremely difficult; second, the correspondences obtained are typically part-wise, which is inadequate for morphing tasks that require dense correspondences. Note that the results within the yellow dashed boxes are from DenseMatcher~\cite{zhu2024densematcher} manuscript.}
    \label{fig: explcit_corres}
\end{figure*}

\subsection{Exploratory Experiments with Explicit Correspondence}
\label{SM: explicit}

Inspired by traditional shape morphing and image morphing, our initial method aimed to establish explicit correspondences between textured 3D representations. Specifically, we sought to assign DIFT~\cite{tang2023emergent} features to each Gaussian~\cite{huang20242d}. This method was based on the fact that DIFT features have been validated to provide 3D correspondence for the same object from different viewpoints and semantic correspondence across objects, making this introduction reasonable.  

However, we overlooked a key issue: the large number of 3D points, which led to two main drawbacks: (a) high computational cost and (b) the feature handling, which works well for single-point-to-single-point searches from a single viewpoint in 2D foundation models, becomes difficult when attempting to preserve the feature's approximate consistency across different viewpoints.  

After extensive optimization and adjustment of parameters, we obtained a mapping and performed texture transfer between two cartoon characters. We found that the learned correspondence was inaccurate and exhibited a "layered" characteristic, which closely resembled the patterns discovered by DenseMatcher~\cite{zhu2024densematcher}. However, this correspondence is unsuitable for morphing and would require substantial research effort to address. Therefore, we are more inclined to pursue morphing research based on promising 3D generation models with implicit correspondence.

\subsection{Testing Protocol and Cases}
\label{SM: testing}

\subsubsection{Quantitative Test Details}

For the FID test, the reference images consist of 3000 samples, which were obtained by rendering 15 sets of 3D pairs from 100 different viewpoints. The test images consist of 1500 samples, generated by each method using the same 3D pairs. The remaining quantitative metrics are obtained from testing on the 15 pairs of data.

When evaluating the structural and semantic plausibility of the generated images using GPT-4o, we input the results generated by six methods on the same 3D pairs into GPT-4o for comparison and scoring (similar to the images in Fig. 6). Meanwhile, we provide a guiding prompt that instructs GPT-4o to engage in a step-by-step reasoning process during the evaluation, enhancing both the interpretability and accuracy of the assessment: \textit{"What I am doing now is morphing with textured 3D representations, the purpose is to generate an intermediate interpolation sequence, and at the same time require the transition from source to target to be smooth and reasonable. Now I have six methods, where the first row will give the source and target, and each of the remaining rows is a method. Columns 1-3 are the first test example (morphing from teapot to bowl), columns 4-6 are the second test example (morphing from polar bear to wooden stool), and columns 7-9 are the third test example (morphing from pumpkin to mushroom). Please help me score these methods for the generated intermediate morphing results in terms of shape rationality and semantic rationality, 0 is the lowest score and 1 is the highest score, and give the discrimination results."}

\subsubsection{Prompts for Testing Cases}

The strength of our method lies in its ability to perform morphing on diverse cross-category or same-category 3D object pairs. This capability was carefully considered during the selection of test cases, which were primarily categorized into furniture, vehicles, plants, humanoid objects, and animals. Moreover, our method goes beyond shape-only morphing, as we also aim to validate its effectiveness on diverse and richly textured color variations. As such, the color range in our test cases is intentionally broad to ensure comprehensive evaluation. More details about the test cases and their corresponding prompts can be found in Tab.~\ref{tab: case}.

\subsection{Raw Statistics of User Study}
\label{SM: user study}

We recruited over 20 volunteers for a user study, where they were asked to rank morphing sequences generated by six methods for the same pairs of test samples. Presenting the results of different methods simultaneously allows users to clearly observe their differences, making the comparison more fair. The 3D object pairs and results are presented in Tab.~\ref{tab: raw_1} and Tab.~\ref{tab: raw_2}.

\begin{table*}[t]
\footnotesize
\centering
\begin{tabular}{ccccccccccccc}
\toprule
& \multicolumn{2}{c}{Teapot - Bowl} 
& \multicolumn{2}{c}{Polar Bear - Stool} 
& \multicolumn{2}{c}{Pumpkin - Mushroom} 
& \multicolumn{2}{c}{Teddy Bear - Boot} 
& \multicolumn{2}{c}{Mario - Luigi} 
& \multicolumn{2}{c}{Average} 

 \\ \cmidrule(lr){2-3} \cmidrule(lr){4-5} \cmidrule(lr){6-7} \cmidrule(lr){8-9}\cmidrule(lr){10-11} \cmidrule(lr){12-13} 
 & STP-U$\uparrow$ & SEP-U$\uparrow$ & STP-U$\uparrow$ & SEP-U$\uparrow$ & STP-U$\uparrow$ & SEP-U$\uparrow$ & STP-U$\uparrow$ & SEP-U$\uparrow$ & STP-U$\uparrow$ & SEP-U$\uparrow$& STP-U$\uparrow$ & SEP-U$\uparrow$\\
\midrule
DiffMorpher& 0.25 & 0.20 & 0.13 & 0.33 & 0.60 & 0.50 & 0.75 & 0.40 & 0.70 & 0.30&0.49&0.35\\ 
AID& 0.15 & 0.53 & 0.40 & 0.27 & 0.50 & 0.60 & 0.25 & 0.32 & 0.70 & 0.60&0.40&0.46\\ 
MV-Adapter& 0.65 & 0.67 & 0.07 & 0.40 & 0.10 & 0.50 & 0.35 & 0.20 & 0.02 & 0.45&0.24&0.44\\ 
Luma&0.50  & 0.13 & 0.73 & 0.60 & 0.10 & 0.25 & 0.60 & 0.60 & 0.00 & 0.25&0.39&0.25\\ 
\rowcolor[HTML]{FFF2CC} 
MorphFlow& 0.45 & 0.47 & 0.67 & 0.53 & 0.70 & 0.40 & 0.30 & 0.52 & 0.40 & 0.40&0.50&0.46\\ 
\rowcolor[HTML]{FFF2CC} 
Ours & \textbf{1.00} & \textbf{1.00} & \textbf{0.40} & \textbf{0.87} &\textbf{ 1.00} & \textbf{0.75} & \textbf{0.75} & \textbf{0.96} &\textbf{ 1.00} & \textbf{1.00} &\textbf{0.83}&\textbf{0.92}\\ 
\bottomrule
\end{tabular}
\caption{The raw statistics for user study (Part 1).}
\label{tab: raw_1}
\end{table*}

\begin{table*}[t]
\footnotesize
\centering
\begin{tabular}{ccccccccccccc}
\toprule
& \multicolumn{2}{c}{Animal Skull - Cow} 
& \multicolumn{2}{c}{Car - Truck} 
& \multicolumn{2}{c}{House - Church} 
& \multicolumn{2}{c}{Chair - Donut} 
& \multicolumn{2}{c}{Tank - Cannon} 
& \multicolumn{2}{c}{Average} 
 \\ \cmidrule(lr){2-3} \cmidrule(lr){4-5} \cmidrule(lr){6-7} \cmidrule(lr){8-9}\cmidrule(lr){10-11} \cmidrule(lr){12-13} 
 & STP-U$\uparrow$ & SEP-U$\uparrow$ & STP-U$\uparrow$ & SEP-U$\uparrow$ & STP-U$\uparrow$ & SEP-U$\uparrow$ & STP-U$\uparrow$ & SEP-U$\uparrow$ & STP-U$\uparrow$ & SEP-U$\uparrow$& STP-U$\uparrow$ & SEP-U$\uparrow$\\
\midrule
DiffMorpher& 0.20 & 0.20 & 0.36 & 0.40 & 0.20 & 0.25 & 0.45 & 0.40 & 0.70 & 0.00&0.38&0.25 \\ 
AID& 0.50 & 0.40 & 0.28 & 0.55 & 0.30 & 0.65 & 0.30 & 0.80 & 0.40 & 0.33&0.36&0.55\\ 
MV-Adapter& 0.20 & 0.27 & 0.28 & 0.40 & 0.10 & 0.10 & 0.45 & 0.20 & 0.00 & 0.33&0.21&0.26\\ 
Luma& 0.30 & 0.60 & 0.52 & 0.45 & 0.70 & 0.35 & 0.20 & 0.00 & 0.50 & 0.67&0.44&0.41\\ 
\rowcolor[HTML]{FFF2CC} 
MorphFlow& 0.80 & 0.53 & 0.56 & 0.30 & 0.70 & 0.65 & 0.60 & 0.60 & 0.40 & 0.67&0.61&0.55\\ 
\rowcolor[HTML]{FFF2CC} 
Ours & \textbf{1.00} & \textbf{1.00} & \textbf{1.00} & \textbf{0.90} & \textbf{1.00} & \textbf{1.00} & \textbf{1.00} & \textbf{1.00} & \textbf{1.00} & \textbf{1.00} &\textbf{1.00}&\textbf{0.98}\\ 
\bottomrule
\end{tabular}
\caption{The raw statistics for user study (Part 2).}
\label{tab: raw_2}
\end{table*}

\begin{table*}[t]
\footnotesize
\centering
\begin{tabular}{ccp{15cm}}
\toprule
Index & Objects & Prompts \\ \midrule
1 & Stool & \textit{"A wooden tripod stool."} \\
2 & Chair & \textit{"A blue plastic chair."} \\
3 & Llama & \textit{"A realistic 3D model of a llama."} \\
4 & Dog & \textit{"A realistic 3D model of a Husky dog with a big head"} \\
5 & Pumpkin & \textit{"A flat, orange, pixelated Lego pumpkin with a green stem."} \\
6 & Mushroom & \textit{"A light green mushroom."} \\
7 & Car & \textit{"A sleek car with smooth curves, shiny metallic surface, and detailed wheels."} \\
8 & Truck & \textit{"A large red truck with a spacious cargo bed, sturdy wheels, and a robust front grille."} \\
9 & Lounge Sofa & \textit{"A purple lounge sofa."} \\
10 & Massage Sofa & \textit{"A pink medieval-style massage sofa with intricate carvings, plush upholstery, and a comfortable, luxurious design."} \\
11 & Mario & \textit{"A cartoon-style Mario character with a red hat, blue overalls, white gloves, and a cheerful expression."} \\
12 & Luigi & \textit{"A cartoon-style Luigi character with a green hat, blue overalls, and a tall, thin build."} \\
13 & Tank & \textit{"A 3D model of a military tank with detailed textures."} \\
14 & Teapot & \textit{"A classic teapot."} \\
15 & Bowl & \textit{"A simple teal bowl."} \\
16 & Fighter Jet & \textit{"A sleek fighter jet with sharp aerodynamic lines, detailed metallic surface, and camouflage paint."} \\
17 & Seagull & \textit{"A seagull with detailed wings, a sleek body, and a realistic beak, in natural white and gray colors."} \\
18 & Cannon & \textit{"A cannon with a long, cylindrical barrel mounted on a wooden carriage."} \\
19 & House & \textit{"A toy house in a fairy-tale style with whimsical architecture, pastel colors, and charming details like a crooked chimney and flower decorations."} \\
20 & Church & \textit{"A classic church with tall spires, arched windows, and a large central entrance."} \\
21 & Skull & \textit{"A 3D model of a human skull."} \\
22 & Animal Skull & \textit{"A 3D low poly model of an animal skull with gray appearance."} \\
23 & Dinosaur & \textit{"A dinosaur with a large, muscular body, a long tail, and realistic skin texture."} \\
24 & Teddy Bear & \textit{"A red teddy bear."} \\
25 & Polar Bear &\textit{ "A low poly model of a polar bear with simplified geometric shapes and flat surfaces, featuring a white body and strong build."} \\
26 & Cow & \textit{"A simple cow model with a stocky body, short legs, and small horns."} \\
27 & Cap & \textit{"A blue baseball cap."} \\
28 & Helmet & \textit{"A medieval helmet with a rounded metal shell and a faceplate."} \\
29 & Donut & \textit{"A donut with a round shape, a hole in the center, and a sugary glaze."} \\
30 & Ice Cream & \textit{"Pink ice cream with a creamy texture, served in a cone or cup."} \\
31 & Hydrant & \textit{"A fire hydrant with a cylindrical body, typically painted in bright red."} \\
32 & Drum & \textit{"A metal oil drum with a cylindrical shape, a top and bottom lid, and a rugged surface." }\\
33 & Vase & \textit{"A light purple vase."} \\
34 & Boot & \textit{"A snow boot with a thick, insulated lining, waterproof exterior, and durable sole for winter conditions."} \\
35 & Fish & \textit{"A chubby fish with a vibrant green body."} \\
\bottomrule
\end{tabular}
\caption{Prompts for testing cases.}
\label{tab: case}
\end{table*}

\end{document}